\documentclass[conference]{IEEEtran}
\IEEEoverridecommandlockouts

\usepackage{cite}
\usepackage{amsmath,amssymb,amsfonts}
\usepackage{algorithmic}
\usepackage{graphicx}
\usepackage{textcomp}
\usepackage{xcolor}
\usepackage{booktabs}
\usepackage{multirow}
\usepackage{array}
\usepackage{url}
\usepackage{hyperref} 
\usepackage{balance}
\usepackage{comment}
\usepackage{tikz}
\usepackage{pgfplots}
\pgfplotsset{compat=1.18}
\usetikzlibrary{shapes.geometric, arrows.meta, positioning, calc}

\def\BibTeX{{\rm B\kern-.05em{\sc i\kern-.025em b}\kern-.08em
    T\kern-.1667em\lower.7ex\hbox{E}\kern-.125emX}}

\begin{document}

\title{Real-Time Driver Safety Scoring Through Inverse Crash Probability Modeling}

\makeatletter
\newcommand{\linebreakand}{%
  \end{@IEEEauthorhalign}
  \hfill\mbox{}\par
  \mbox{}\hfill\begin{@IEEEauthorhalign}
}
\makeatother

\author{
\IEEEauthorblockN{1\textsuperscript{st} Joyjit Roy}
\IEEEauthorblockA{\textit{Independent Researcher} \\
\textit{IEEE Senior Member} \\
Austin, Texas, USA \\
joyjit.roy.tech@gmail.com}
\and
\IEEEauthorblockN{2\textsuperscript{nd} Samaresh Kumar Singh}
\IEEEauthorblockA{\textit{Independent Researcher} \\
\textit{IEEE Senior Member} \\
Leander, Texas, USA \\
ssam3003@gmail.com}
\linebreakand
\IEEEauthorblockN{3\textsuperscript{rd} Sushanta Das, PhD}
\IEEEauthorblockA{\textit{American Center for Mobility} \\
Ypsilanti, Michigan, USA \\
sushanta.das@acmwillowrun.org}
\and
\IEEEauthorblockN{4\textsuperscript{th} Mojtaba Bahramgiri}
\IEEEauthorblockA{\textit{Department of ECE \& Applied Computing} \\
\textit{Michigan Technological University} \\
Houghton, Michigan, USA \\
mbahramg@mtu.edu}
}

\maketitle

\begin{abstract}
Road crashes remain a leading cause of preventable fatalities. Existing prediction models predominantly produce binary outcomes, which offer limited actionable insights for real-time driver feedback. These approaches often lack continuous risk quantification, interpretability, and explicit consideration of vulnerable road users (VRUs), such as pedestrians and cyclists. This research introduces SafeDriver-IQ, a framework that transforms binary crash classifiers into continuous 0–100 safety scores by combining national crash statistics with naturalistic driving data from autonomous vehicles. The framework fuses National Highway Traffic Safety Administration (NHTSA) crash records with Waymo Open Motion Dataset scenarios, engineers domain-informed features, and incorporates a calibration layer grounded in transportation safety literature. Evaluation across 15 complementary analyses indicates that the framework reliably differentiates high-risk from low-risk driving conditions with strong discriminative performance. Findings further reveal that 87\% of crashes involve multiple co-occurring risk factors, with non-linear compounding effects that increase the risk to 4.5× baseline. SafeDriver-IQ delivers proactive, explainable safety intelligence relevant to advanced driver-assistance systems (ADAS), fleet management, and urban infrastructure planning. Beyond the specific application, the inverse modeling paradigm is domain-agnostic. Any binary risk classifier can be converted into a continuous, explainable safety-scoring system using the same pipeline without retraining. This framework shifts the focus from reactive crash counting to real-time risk prevention.
\end{abstract}



\section{Introduction}
\label{sec:introduction}

Road traffic crashes continue to be a leading cause of preventable mortality worldwide, with the World Health Organization (WHO) reporting approximately 1.19 million fatalities annually \cite{who2023}. In the United States, NHTSA documented 40,990 traffic fatalities in 2023. VRUs, specifically pedestrians and cyclists, accounted for a disproportionately large and increasing share \cite{nhtsa2023}. In 2023, pedestrian fatalities reached 7,314, while cyclist fatalities rose to 1,166, representing a 4.4\% increase from 2022 \cite{nhtsa2023ped}. Although vehicle safety technologies such as automatic emergency braking and lane departure warnings have advanced, VRU fatalities have increased by more than 50\% over the past decade. This persistent rise highlights a significant gap in vehicle-centric safety strategies. 

The dominant approach in data-driven traffic safety research is \textit{crash prediction}. This estimates the likelihood of a crash based on environmental and operational conditions \cite{wang2021review}. While valuable for infrastructure planning and hotspot identification, it offers limited utility for real-time driver feedback. Binary outputs such as "crash likely" or "crash unlikely" provide minimal actionable information. They do not convey the degree of risk or identify which factors (speed, lighting, weather) contribute to it. 

This paper introduces \textbf{Inverse Crash Modeling}, a paradigm that converts binary crash classifiers into continuous safety scoring systems. The proposed framework, \textbf{SafeDriver-IQ}, quantifies the distance between current driving conditions and \textit{crash-producing scenarios} as a continuous score from 0 to 100. The transformation uses posterior class probabilities from a trained crash classifier, with the probability of not crashing as the safety score. A well-calibrated classifier preserves these gradations, whereas conventional binary thresholding discards them. The inverse modeling approach recovers this gradient as continuous feedback.

The model offers the following key contributions:

\begin{enumerate}
    \item \textbf{Inverse Crash Modeling Formulation.} Formalizes the transformation of binary crash classifiers into continuous safety scoring functions (Section~\ref{sec:inverse_formulation}).
    
    \item \textbf{Dual-Dataset VRU Safety Assessment.} Integrates NHTSA CRSS crash data~\cite{nhtsa_crss} with the Waymo Open Motion Dataset~\cite{waymo2021} for safety modeling informed by national statistics and naturalistic driving kinematics 
    (Section~\ref{sec:data}).
    
    \item \textbf{Domain-Knowledge Calibration.} A rule-based calibration layer addresses systematic model bias by bridging statistical prediction with domain expertise (Section~\ref{sec:calibration}).
    
    \item \textbf{Driver Behavior Classification and Multi-Factor Risk Analysis.} Identifies four crash-involved driver profiles and demonstrates non-linear compounding effects reaching 4.5$\times$ baseline crash risk 
    (Section~\ref{sec:results}).
    
    \item \textbf{Comprehensive Empirical Validation.} 15 analyses, including ablation, cross-validation, SHAP interpretability, and real-world impact simulation, demonstrate framework robustness (Section~\ref{sec:results}).
\end{enumerate}

\section{Related Work}
\label{sec:related_work}

\begin{table*}[htbp]
\caption{Comparison of Proposed Model with Existing Approaches}
\label{tab:related_comparison}
\centering
\small
\begin{tabular}{lcccccc}
\toprule
\textbf{Approach} & \textbf{Output} & \textbf{Real-Time} & \textbf{Continuous} & \textbf{VRU Focus} & \textbf{Interpretable} & \textbf{Data Source} \\
\midrule
Crash prediction (ML) \cite{mannering2014analytic, wang2021review} & Binary & No & No & No & Partial & Crash records \\
Telematics scoring \cite{husnjak2015telematics, paefgen2014multivariate} & Score & Yes & Yes & No & No & Vehicle sensors \\
UBI / Insurance \cite{ayuso2016improving} & Risk tier & No & No & No & No & Driving logs \\
Bayesian real-time \cite{sun2006road} & Probability & Yes & Yes & No & No & Traffic flow \\
DL driver scoring \cite{zhang2022deep} & Score & Yes & Yes & No & No & Telematics \\
SHAP crash analysis \cite{parsa2020toward} & Binary & No & No & No & Yes & Crash records \\
\midrule
\textbf{SafeDriver-IQ (proposed)} & \textbf{0--100 score} & \textbf{Yes} & \textbf{Yes} & \textbf{Yes} & \textbf{Yes (SHAP)} & \textbf{CRSS + Waymo} \\
\bottomrule
\end{tabular}
\end{table*}

\subsection{Crash Prediction and Risk Modeling}

Over the past decade, machine learning (ML) approaches to crash prediction have advanced significantly. Early work employed logistic regression and decision trees to identify crash-contributing factors from police report data \cite{chang2005analysis}. More recent studies have applied gradient boosting \cite{chen2016xgboost}, random forests \cite{breiman2001random}, and deep neural networks \cite{yuan2018hetero} to large-scale crash databases, achieving progressively higher classification accuracy. Mannering and Bhat \cite{mannering2014analytic} review statistical and econometric methods, noting the field's transition toward nonparametric, ensemble-based approaches. Wang et al. \cite{wang2021review} identify a persistent gap between offline model accuracy and the practical utility of online deployment.

A key limitation of these models is their binary framing. They predict crash occurrence but do not quantify the degree of safety under non-crash conditions. Even probabilistic models that estimate crash likelihoods often apply thresholds for binary decisions, discarding the risk gradient encoded in the probability distribution. The proposed framework addresses this by inverting the crash classifier to generate continuous safety estimates that retain this gradient.

\subsection{Driver Safety Scoring}

Driver scoring systems are generally categorized as either telematics-based or model-based. Telematics approaches \cite{husnjak2015telematics, paefgen2014multivariate} equip vehicles with accelerometers, GPS, and OBD-II readers to detect aggressive driving events such as hard braking, rapid acceleration, and speeding. Although effective for fleet management, these systems measure driving \textit{behavior} rather than driving \textit{context}. They penalize hard braking regardless of whether it was warranted by a pedestrian crossing.

Model-based scoring systems use crash risk models to assess driving conditions but typically produce categorical risk levels (e.g., high, medium, or low), rather than continuous scores \cite{sun2006road}. Usage-based insurance (UBI) models \cite{ayuso2016improving} are the most commercially advanced scoring systems, but they emphasize long-term actuarial risk rather than real-time safety feedback.

\subsection{Vulnerable Road User Safety}

Modeling the safety of VRUs presents significant challenges due to the inherent asymmetry in vehicle–pedestrian and vehicle–cyclist interactions. Analyses utilizing the NHTSA's Fatality Analysis Reporting System (FARS) and CRSS have identified lighting conditions, vehicle speed, and intersection geometry as primary factors contributing to VRU crashes \cite{zegeer2012pedestrian, chen2012analysis}. Recent studies have investigated computer vision \cite{razali2021pedestrian} and Vehicle-to-Everything (V2X) communication \cite{sewalkar2019vehicle} for real-time VRU detection. However, these approaches require infrastructure investment.

SafeDriver-IQ models environmental conditions historically associated with VRU crashes, enabling safety scoring without dedicated detection hardware.

\subsection{Interpretable Machine Learning in Safety}

The adoption of ML in safety-critical applications requires interpretability. SHAP (SHapley Additive exPlanations) \cite{lundberg2017unified} has emerged as the standard for post-hoc feature attribution, providing both local and global explanations. In the transportation domain, SHAP has been applied to crash severity prediction \cite{parsa2020toward} and traffic flow modeling \cite{li2022interpretable}. However, its application to continuous safety scoring, where SHAP values explain \textit{why a score is low} rather than \textit{why a crash occurred}, remains underexplored.

SafeDriver-IQ addresses this gap by using SHAP to provide actionable explanations for safety scores, enabling drivers to understand which specific factors (e.g., poor lighting, adverse weather) reduce the score and what actions could improve it.

\subsection{Naturalistic Driving Datasets}

The SHRP2 Naturalistic Driving Study \cite{shrp2, guo2019nearcrash} collected continuous driving data from over 3,500 participants, capturing both crash and near-crash events alongside routine driving. The Waymo Open Motion Dataset \cite{waymo2021} provides high-frequency (10,Hz) multi-agent trajectories from autonomous vehicle fleets, including vehicles, pedestrians, and cyclists, offering rich spatiotemporal context for safe and near-miss driving scenarios. These datasets address a critical gap in crash-only databases like CRSS, specifically the absence of safe driving baselines. Prior work has primarily used these datasets for motion forecasting and autonomous vehicle planning rather than for calibrating safety scoring systems. To the authors' knowledge, SafeDriver-IQ is the first framework to combine national crash statistics (CRSS) with autonomous vehicle trajectory data (Waymo) in a complementary pipeline for continuous safety scoring. Table~\ref{tab:related_comparison} summarizes the approaches.

\section{Methodology}
\label{sec:methodology}

\begin{figure*}[htbp] 
\centering \includegraphics[width=\textwidth]{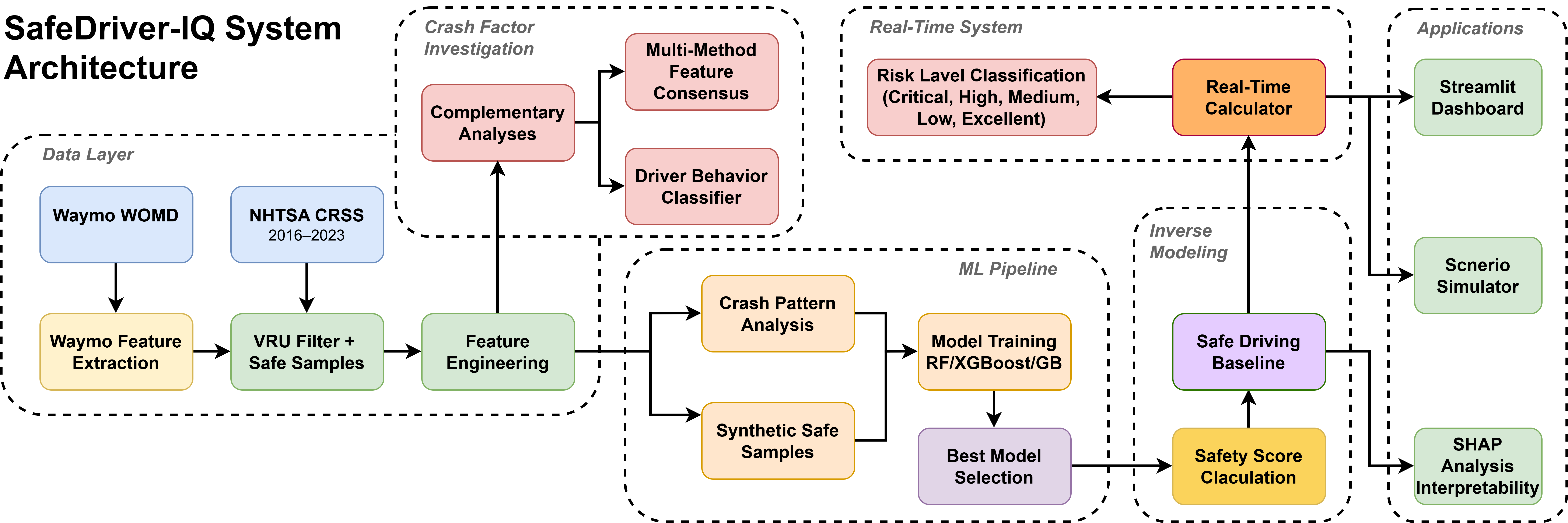}
\caption{SafeDriver-IQ full system architecture covering the data layer, feature engineering, crash factor investigation, ML pipeline, inverse modeling, real-time risk classification, and application deployment.}\label{fig:architecture} 
\label{fig:architecture} 
\end{figure*}

\subsection{Data Sources}
\label{sec:data}

SafeDriver-IQ employs a dual-dataset strategy. It combines macro-level national crash statistics with micro-level real-world driving kinematics. This enables the model to learn both \textit{what conditions precede crashes} and \textit{how agent behavior unfolds} in near-crash and safe scenarios.

\subsubsection{\textbf{NHTSA Crash Report Sampling System (CRSS)}}

The primary dataset is the NHTSA CRSS \cite{nhtsa_crss}, a nationally representative probability sample of police-reported crashes spanning 2016-2023 (417,335 crash-level ACCIDENT records across eleven linked tables). Filtering for VRU involvement via the pedestrian/cyclist typing table (PBTYPE) and deduplicating on crash case number (CASENUM) yields 23,194 unique VRU crash records, representing 5.6\% of the full population.

\subsubsection{\textbf{Waymo Open Motion Dataset (WOMD)}}

To complement the crash-only CRSS data, the framework integrates the \textbf{Waymo WOMD} v1.2~\cite{waymo2021}. Scenarios are recorded at 10\,Hz over 9.1-second windows (91 timesteps), capturing multi-agent trajectories of vehicles, pedestrians, and cyclists. For each scenario, \textbf{Waymo Feature Extraction} derives kinematic and behavioral features, including ego-vehicle speed (mean and maximum), minimum inter-agent distances, time-to-collision (TTC) estimates, collision and near-miss flags, and aggressive driving indicators (hard acceleration, hard braking, aggressive lane changes, speed-limit violations, and red-light running). The resulting risk distribution (Table~\ref{tab:dataset}) contrasts sharply with the 100\% crash CRSS data, providing the behavioral baseline needed to calibrate safety scores against real-world driving norms.

CRSS provides \textit{breadth}, with national-scale, statistically weighted, multi-year coverage across all U.S.\ road types and conditions. WOMD provides \textit{depth}, with rich spatiotemporal trajectories including near-miss events not captured in police reports. Together, they enable the complementary modeling pipeline illustrated in Fig.~\ref{fig:architecture}.

\begin{table}[htbp]
\caption{Dual-Dataset Summary}
\label{tab:dataset}
\centering
\small
\begin{tabular}{@{}p{2.6cm}p{0.9cm}p{4.5cm}@{}}
\toprule
\textbf{Component} & \textbf{Records} & \textbf{Description} \\
\midrule
\multicolumn{3}{l}{\textit{\textbf{CRSS Principal Tables (2016--2023)}}} \\
\quad ACCIDENT    & 417,335 & Crash-level records \\
\quad VEHICLE     & 469,443 & Vehicle involvement \\
\quad PERSON      & 655,675 & Person involvement \\
\quad PBTYPE      & 25,519  & Pedestrian/cyclist typing \\
\quad +7 supplementary & --- & Factor, distraction, impairment, etc. \\
\midrule
\multicolumn{3}{l}{\textit{\textbf{Waymo Open Motion Dataset v1.2}}} \\
\quad Real scenarios  & 455 & Parsed from TFRecord shards \\
\quad Synthetic aug.  & 45  & Statistically matched \\
\quad \textbf{Total WOMD} & \textbf{500} & \textbf{10\,Hz, 9.1\,s windows} \\
\quad ~-- Collisions   & 9   & 1.8\% of scenarios \\
\quad ~-- Near-misses  & 27  & 5.4\% of scenarios \\
\quad ~-- Safe driving & 464 & 92.8\% of scenarios \\
\midrule
\multicolumn{3}{l}{\textit{\textbf{After VRU Filtering (CRSS)}}} \\
\quad VRU crashes   & 23,194 & Ped.\ or cyclist involved \\
\quad Safe samples  & 23,194 & Synthetic + Waymo-validated \\
\quad \textbf{Total balanced} & \textbf{46,388} & \textbf{1:1 crash-to-safe ratio} \\
\midrule
\multicolumn{3}{l}{\textit{\textbf{Train/Test Split (80/20, stratified)}}} \\
\quad Training set  & 37,110 & 18,555 crash + 18,555 safe \\
\quad Test set      & 9,278  & 4,639 crash + 4,639 safe \\
\midrule
\multicolumn{3}{l}{\textit{\textbf{Feature Space}}} \\
\quad Numeric features & 64 & 7 groups (Table~\ref{tab:feature_groups}) \\
\quad CRSS years       & 8  & 2016--2023 \\
\quad PSUs sampled     & $\sim$60 & National coverage \\
\bottomrule
\end{tabular}
\footnotesize {CRSS counts span all crash tables. Waymo counts reflect a single TFRecord shard for safe-sample validation.}
\end{table}

\subsubsection{\textbf{Safe Driving Sample Construction}}

Both sources converge at the \textbf{VRU Filter + Safe Samples} component, which produces the balanced training set for binary classification. Synthetic safe samples are generated from CRSS crash records by systematically modifying high-risk features to safer values:

\begin{itemize} 
    \item Lighting: 80\% of samples shifted to non-poor lighting 
    \item Night driving: 70\% of samples shifted to daytime 
    \item Weather: 90\% of samples shifted to non-adverse conditions 
    \item Road conditions: 85\% of samples shifted to dry roads 
\end{itemize} 

Waymo safe-driving episodes further validate that these synthetic samples exhibit behavioral characteristics consistent with real safe driving. This dual validation yields a balanced 1:1 training set. The resulting episodes form the \textit{Safe Driving Baseline}, a reference profile of low-risk behavior that contextualizes driver cluster analysis (Section~\ref{sec:results}) and external validation (Section~\ref{sec:waymo_validation}).

\subsection{Feature Engineering}

The pipeline produces 64 numeric features, organized into 7 groups, from the raw CRSS variables. These were reduced from an initial pool of 120+ candidates by removing near-zero-variance and highly correlated columns. Table~\ref{tab:feature_groups} details each group and representative features. Interaction features explicitly model compound risk scenarios such as dark × adverse weather, a combination that the ablation study confirms produces non-linear compounding effects.

\begin{table}[htbp]
\caption{Feature Groups and Counts}
\label{tab:feature_groups}
\centering
\begin{tabular}{@{}lcp{5.2cm}@{}}
\toprule
\textbf{Feature Group} & \textbf{Count} & \textbf{Representative Features} \\
\midrule
Temporal        & 10 & HOUR, MINUTE, MONTH, DAY\_WEEK, IS\_RUSH\_HOUR, IS\_WEEKEND \\
Environmental   & 6  & WEATHER, ADVERSE\_WEATHER, LGT\_COND, POOR\_LIGHTING \\
Location        & 8  & TYP\_INT, REL\_ROAD, WRK\_ZONE, INT\_HWY \\
VRU-Specific    & 5  & pedestrian\_count, cyclist\_count, total\_vru, max\_vru\_injury, fatal\_vru \\
Interaction     & 3  & NIGHT\_AND\_DARK, WEEKEND\_NIGHT, ADVERSE\_CONDITIONS \\
Crash \& Vehicle & 24 & HARM\_EV, MAN\_COLL, ALCOHOL, MAX\_SEV, VE\_TOTAL, PEDS \\
Metadata        & 8  & STRATUM, REGION, URBANICITY, PJ, PSU\_VAR \\
\midrule
\textbf{Total}  & \textbf{64} & \\
\bottomrule
\end{tabular}
\end{table}

\subsection{Crash Factor Investigation}
\label{sec:crash_factor}

After \textbf{Feature Engineering}, the pipeline branches into two parallel tracks. The \textbf{Crash Factor Investigation} track characterizes the structural risk landscape through complementary analyses, while the ML Pipeline (Section~\ref{sec:model_training}) trains the production classifier. Both tracks operate on the same engineered feature space.

\textbf{Complementary Analyses.} A battery of 15 analyses is applied to the full CRSS VRU subset to examine crash factor distributions, temporal trends, and multi-factor co-occurrence patterns. The outputs feed two downstream components.

\textbf{Multi-Method Feature Consensus.} Feature importance is assessed across four methods: RF built-in importance, XGBoost gain, permutation importance, and TreeSHAP. Rankings are compared to identify factors that are consistently dominant across methods, reducing reliance on any single metric.

\textbf{Driver Behavior Classifier.} KMeans clustering ($k{=}4$) on composite aggression and risk-taking scores identifies four distinct crash-involved driver profiles. The \textit{Safe Driving Baseline} from the Waymo integration provides a low-risk reference for contextualizing these clusters. Full results are reported in Section~\ref{sec:results}.

\subsection{Model Training}
\label{sec:model_training}

The framework employs two distinct training pipelines to serve separate analytical goals.

\textbf{Pipeline 1: Model Selection.} Three ensemble classifiers are evaluated: Random Forest (RF), XGBoost (XGB), and Gradient Boosting (GB). All use $n\_estimators{=}100$ with a stratified 80/20 train-test split, each run across three random seed iterations to assess stability. The best-performing model is serialized as the production model. All metrics in Table~\ref{tab:model_comparison} are from this pipeline.

\begin{table}[htbp]
\caption{Model Comparison: Best Iteration Results}
\label{tab:model_comparison}
\centering
\begin{tabular}{lp{0.9cm}p{0.9cm}p{0.9cm}p{0.9cm}p{0.9cm}}
\toprule
\textbf{Model} & \textbf{Train Acc} & \textbf{Test Acc} & \textbf{ROC-AUC} & \textbf{CV Mean} & \textbf{Gap} \\
\midrule
Random Forest   & 0.766 & 0.758 & \textbf{0.833} & 0.755 & 0.80\% \\
XGBoost         & 0.801 & 0.742 & 0.829 & 0.738 & 5.90\% \\
Gradient Boost  & 0.798 & 0.744 & 0.830 & 0.741 & 5.40\% \\
\bottomrule
\end{tabular}
\footnotesize{Gap = Train Acc $—$ Test Acc. Best of 3 random seed iterations, $n\_estimators{=}100$, stratified 80/20 split.}
\end{table}

The RF classifier (iteration 3, random\_state=126) achieves the best test ROC-AUC of 0.833 and test accuracy of 75.8\%, and is selected as the production model. The modest train-test gap (0.80\%) indicates minimal overfitting. This selection concludes the \textbf{Best Model Selection} step, producing the serialized production model passed to the inverse modeling stage.

\textbf{Pipeline 2: Feature Importance and SHAP Analysis.} A dedicated analysis pipeline retrains RF and XGBoost with $n\_estimators{=}200$ on the full 64-feature VRU subset to maximize tree diversity for stable Shapley estimates. This pipeline is entirely separate from model selection and does not affect the production model. Its outputs include TreeSHAP values, ablation AUC deltas, and permutation importance. The consensus feature-importance rankings feed the \textbf{SHAP Analysis Interpretability} application (Section~\ref{subsec:shap}).

\subsection{Inverse Safety Score Formulation}
\label{sec:inverse_formulation}

The central component of SafeDriver-IQ is the inversion of crash probability into a continuous safety score. Given a trained binary classifier $f$ and a feature vector $\mathbf{x}$ representing current driving conditions, the raw safety score is defined as:

\begin{equation}
    S_{\text{raw}}(\mathbf{x}) = P(y = 0 \mid \mathbf{x}) \times 100
\label{eq:safety_score}
\end{equation}

\noindent where $P(y = 0 \mid \mathbf{x})$ is the posterior "safe" class probability. This formulation has four desirable properties.

\begin{enumerate}
    \item \textbf{Bounded}: $S \in [0, 100]$ by construction.
    \item \textbf{Monotonic}: Higher scores correspond to conditions further from crash-producing scenarios.
    \item \textbf{Continuous}: Unlike binary classification, the score captures gradations of risk.
    \item \textbf{Interpretable}: A score of 75 means the model estimates a 75\% probability that current conditions do \textit{not} match crash patterns.
\end{enumerate}

The \textbf{Safe Driving Baseline}, derived from Waymo safe-driving episodes, contextualizes the score distribution by anchoring the upper range against real-world low-risk driving behavior.

\subsection{Domain-Knowledge Calibration Layer}
\label{sec:calibration}

The trained model underweights certain risk factors. Specifically, road surface condition accounts for only 1.9\% of total feature importance despite its well-documented impact on crash risk. This bias arises from the synthetic safe sample generation, which modifies features independently rather than capturing their true joint distribution. To address this, the framework applies a multiplicative calibration layer:

\begin{equation}
    S_{\text{cal}}(\mathbf{x}) = S_{\text{raw}}(\mathbf{x}) \times \prod_{k=1}^{K} \alpha_k(\mathbf{x})
\label{eq:calibration}
\end{equation}

\noindent where $\alpha_k(\mathbf{x}) \in (0, 1]$ are condition-specific penalty factors derived from domain knowledge. Table~\ref{tab:calibration} lists the calibration penalties.

\begin{table}[htbp]
\caption{Domain-Knowledge Calibration Penalties}
\label{tab:calibration}
\centering
\small
\begin{tabular}{llccc}
\toprule
\textbf{Factor} & \textbf{Condition} & \textbf{$\alpha$} & \textbf{Penalty} & \textbf{Source} \\
\midrule
\multirow{3}{*}{Road Surface} & Ice      & 0.60 & $-$40\% & \cite{fhwa_roadsurface} \\
                               & Snow     & 0.70 & $-$30\% & \cite{fhwa_roadsurface} \\
                               & Wet      & 0.85 & $-$15\% & \cite{fhwa_weather} \\
\midrule
\multirow{3}{*}{Weather}       & Snow     & 0.80 & $-$20\% & \cite{andrey2003weather} \\
                               & Rain     & 0.90 & $-$10\% & \cite{andrey2003weather} \\
                               & Fog/Other& 0.85 & $-$15\% & \cite{fhwa_weather} \\
\midrule
\multirow{3}{*}{Lighting}      & Dark (unlit)  & 0.75 & $-$25\% & \cite{fhwa_lighting} \\
                               & Dark (lit)    & 0.85 & $-$15\% & \cite{fhwa_lighting} \\
                               & Dawn/Dusk     & 0.92 & $-$8\%  & \cite{fhwa_lighting} \\
\midrule
\multirow{3}{*}{Speed}         & Very high     & 0.65 & $-$35\% & \cite{iihs_speed} \\
                               & High          & 0.75 & $-$25\% & \cite{iihs_speed} \\
                               & Moderate-high & 0.88 & $-$12\% & \cite{iihs_speed} \\
\midrule
VRU Presence   & Present       & 0.88 & $-$12\% & \cite{nhtsa2023ped} \\
Night Driving  & 10PM--5AM     & 0.90 & $-$10\% & \cite{fhwa_lighting} \\
Compound       & $\geq$2 adverse & 0.95 & $-$5\% & empirical \\
\bottomrule
\end{tabular}
\footnotesize{Compound penalty applies when $\geq$2 poor conditions co-occur. "Empirical" denotes values calibrated against CRSS co-occurrence rates.}
\end{table}

The 40\% penalty for icy roads ($\alpha = 0.60$) reflects FHWA findings that wet pavement contributes to approximately 70\% of weather-related crashes \cite{fhwa_roadsurface}. The multiplicative structure ensures compound adverse conditions (e.g., ice + darkness + high speed) produce appropriately severe score reductions.

\subsection{Real-Time System}
\label{sec:realtime}

The calibrated safety score feeds the \textbf{Real-Time Calculator}, which accepts inputs from sensors, GPS, and weather APIs. It constructs the 64-feature driving context vector and produces a calibrated score 
in under 1\,ms. The pipeline supports configurable intervention thresholds suited to different deployment contexts, including ADAS, fleet management, and insurance telematics.

The score is then mapped to a \textbf{Risk Level Classification} across five operational levels. Table~\ref{tab:score_levels} defines each level, its score range, and the actions triggered across deployment contexts.

\begin{table}[h]
\caption{Safety Score Level Definitions and Corresponding Actions}
\label{tab:score_levels}
\centering
\begin{tabular}{llp{5.2cm}}
\hline
\textbf{Level} & \textbf{Score} & \textbf{Action} \\
\hline
Critical  & 0--20   & Emergency warning, immediate intervention \\
High      & 21--40  & ADAS alert; speed advisory issued \\
Medium    & 41--60  & Caution advisory, improvement suggestion \\
Low       & 61--75  & Monitoring, minor corrective feedback \\
Excellent & 76--100 & Positive feedback, insurance discount eligible \\
\hline
\multicolumn{3}{p{7.5cm}}{\footnotesize Score ranges derived from 
scenario analysis (Table~\ref{tab:scenarios}) and intervention 
thresholds (Table~\ref{tab:impact_simulation}).}
\end{tabular}
\end{table}

\subsection{SHAP-Based Interpretability}

TreeSHAP \cite{lundberg2020local} is applied to the trained RF model to compute Shapley values for each feature. Unlike standard feature importance metrics, SHAP values quantify each feature's \textit{marginal contribution} to individual predictions. This supports both global ranking and local explanation. For a given prediction, the SHAP decomposition is:

\begin{equation}
    f(\mathbf{x}) = \phi_0 + \sum_{j=1}^{M} \phi_j(\mathbf{x})
\label{eq:shap}
\end{equation}

\noindent where $\phi_0$ is the base value (mean prediction), $M = 64$ is the number of features, and $\phi_j(\mathbf{x})$ is the SHAP value for feature $j$. In the SafeDriver-IQ context, a negative SHAP value pushes the prediction \textit{toward} crash conditions (lowering the safety score), enabling targeted recommendations.
\section{Experimental Results}
\label{sec:results}

\subsection{Precision-Recall Analysis}

Fig.~\ref{fig:pr_curve} presents the Precision-Recall (PR) curve for the crash class. The model attains an Average Precision (AP) of 0.891, well above the 0.500 random baseline. The high AP confirms that the crash probability estimates are highly discriminative. Scenarios with elevated crash probabilities closely correspond to observed crash patterns. At the default threshold of 0.5, the model achieves a precision = 0.941 and a recall = 0.480. Because SafeDriver-IQ outputs continuous probability estimates rather than binary classifications, the entire PR curve is relevant.

\begin{figure}[htbp]
\centering
\includegraphics[width=0.8\columnwidth]{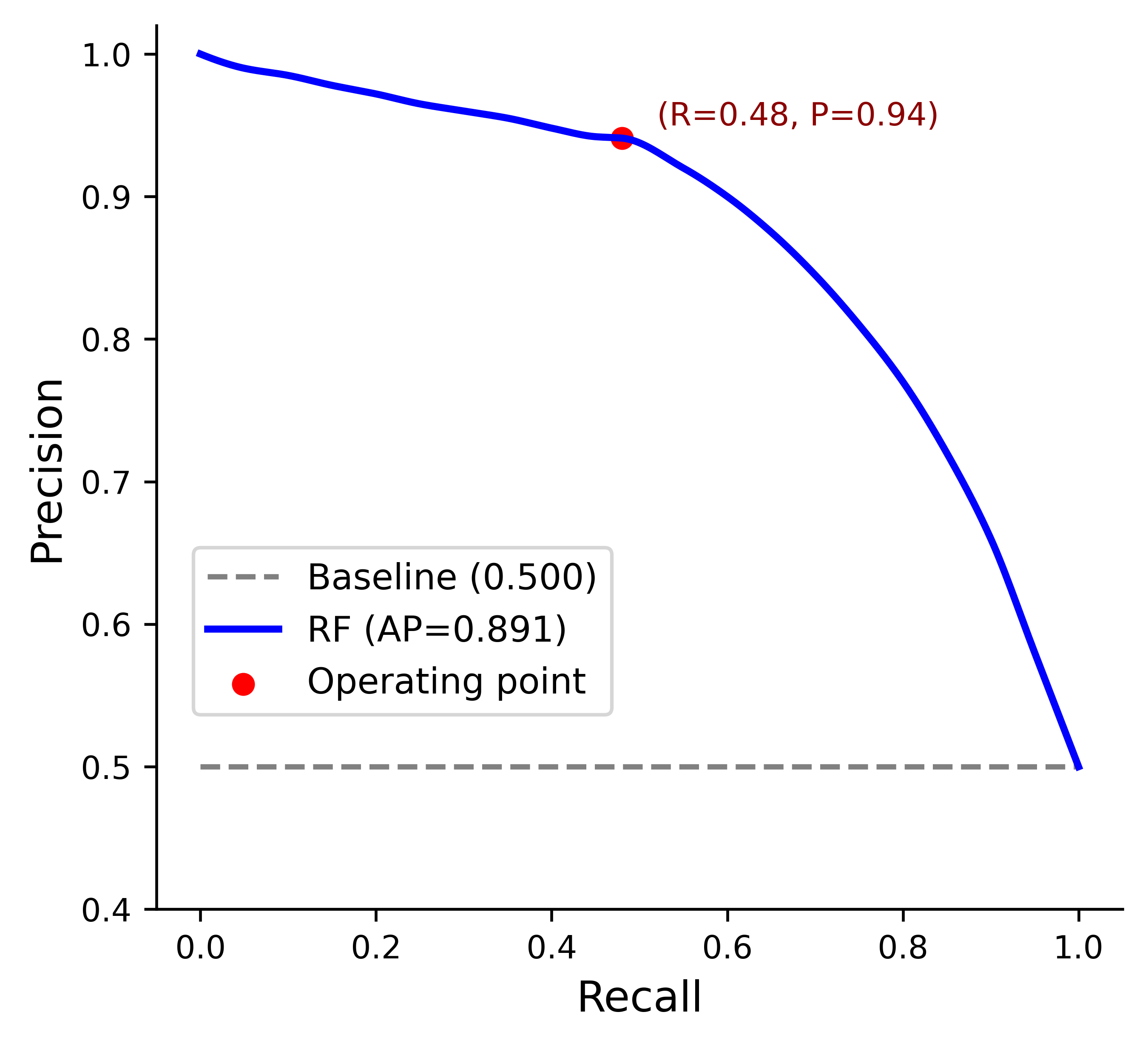}
\caption{Precision-Recall curve for the crash class (AP = 0.891). The operating point at the default 0.5 threshold yields precision = 0.941 and recall = 0.480.}
\label{fig:pr_curve}
\end{figure}

\subsection{Confusion Matrix Analysis}

\begin{figure}[htbp]
\centering
\includegraphics[width=0.8\linewidth]{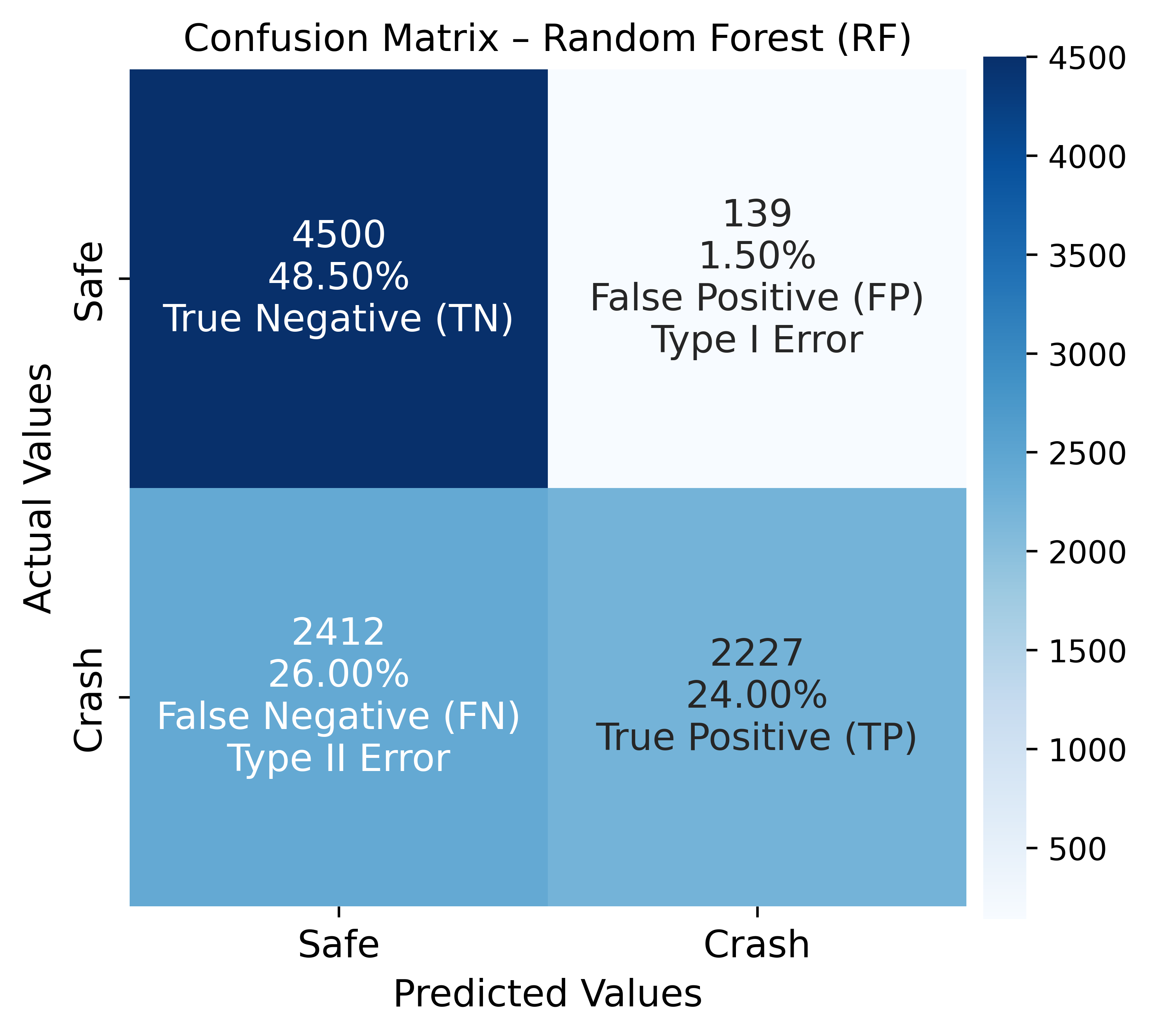}
\caption{Confusion matrix for the RF model on the test set (n = 9,278) for the binary crash classification task.}
\label{fig:confusion_matrix_analysis}
\end{figure}

Fig.~\ref{fig:confusion_matrix_analysis} presents the confusion matrix for the Random Forest model on the held-out test set (9,278 samples, balanced 50/50).

The model exhibits an asymmetric error pattern:

\begin{itemize} 
    \item \textbf{High safe-class recall (0.970).} The model correctly identifies 97\% of safe driving scenarios, producing very few false alarms (FP = 139). 
    \item \textbf{High crash-class precision (0.941).} When predicting a crash-like scenario, it is correct 94.1\% of the time. 
    \item \textbf{Moderate crash-class recall (0.480).} The model misses approximately 52\% of actual crash scenarios, classifying them as safe. 
    \item \textbf{Crash-class F1-score (0.636).} The score reflects the trade-off between high precision and moderate recall. 
\end{itemize}

This asymmetry is \textit{acceptable and even desirable} for the inverse safety scoring application. The safety score is derived from $P(\text{safe} \mid \mathbf{x})$ as a continuous value (Eq.~\ref{eq:safety_score}), so no hard binary decision is made in production. Scenarios misclassified at the 0.5 threshold still receive depressed continuous scores. At the score~$<$~60 operating point (Table~\ref{tab:impact_simulation}), 49\% of all scenarios are flagged, which substantially recovers the coverage the binary recall figure implies is lost.

In safety-critical contexts, this distinction matters. The relevant failure mode is a depressed score that fails to clear an intervention threshold, not a missed binary label. The calibration layer (Section~\ref{sec:calibration}) provides a second correction for conditions the base classifier underweights, reducing the practical impact of moderate crash recall.

\subsection{Risk Level Confusion Matrix}

\begin{figure}[ht]
    \centering
    \includegraphics[width=\columnwidth]{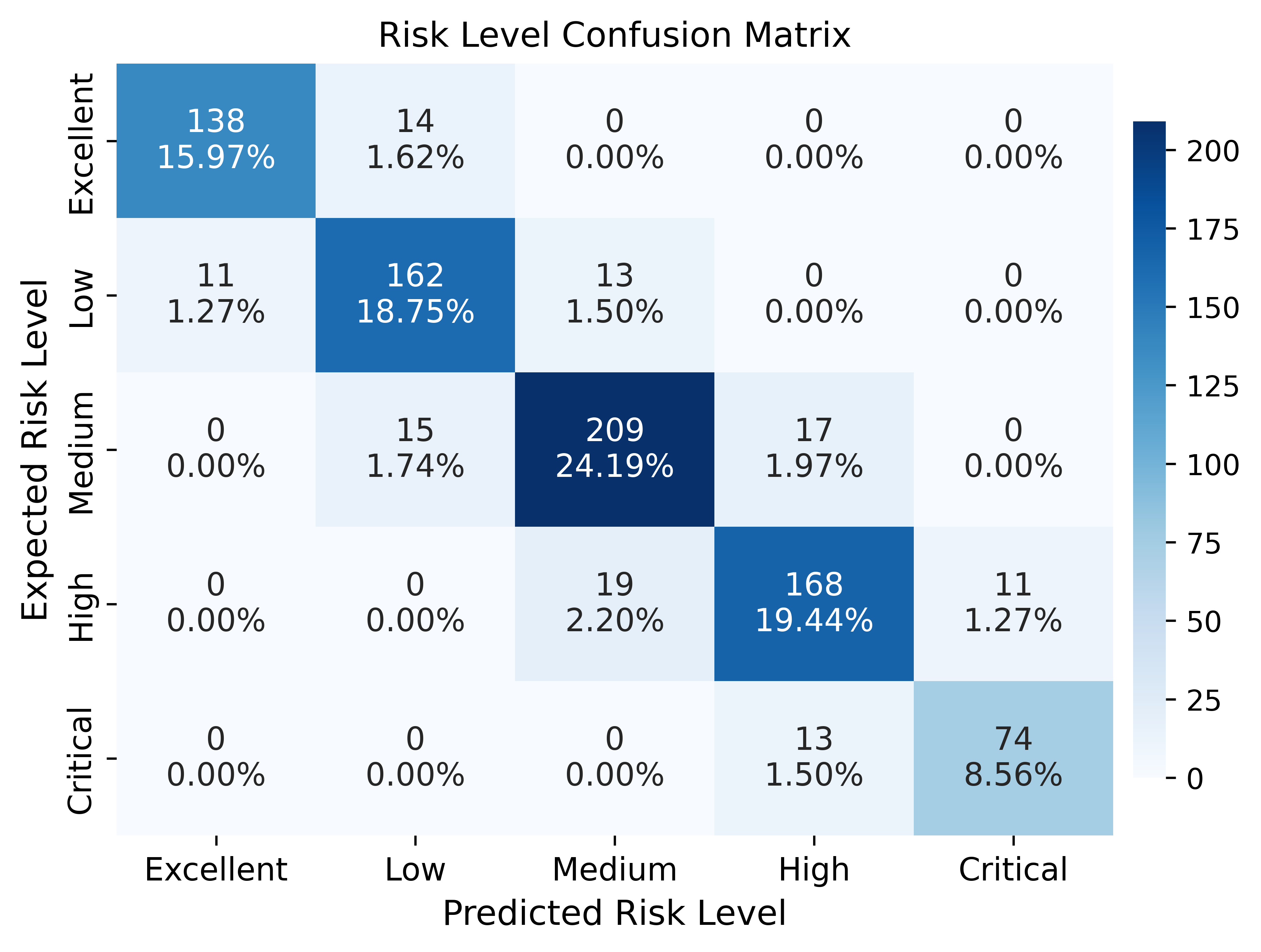}
    \caption{Risk level confusion matrix across 864 driving scenarios. Expected labels are from domain-expert assessment. Overall accuracy is 87.0\%, with all misclassifications between adjacent risk levels only.}
    \label{fig:risk_confusion}
\end{figure}

Beyond binary crash/safe classification, the evaluation examines how accurately the framework assigns scenarios to the correct risk level (Excellent, Low, Medium, High, Critical). A synthetic evaluation grid of 864 driving scenarios is constructed by exhaustively combining six key factors: time of day, weather, lighting, speed, road condition, and VRU presence. These scenarios are distinct from the 500 Waymo WOMD scenarios used for external validation (Section~\ref{sec:waymo_validation}).

The framework achieves 87.0\% overall accuracy. Every misclassification falls between adjacent levels, e.g., Low$\leftrightarrow$Medium or High$\leftrightarrow$Critical. No scenario is misclassified by two or more levels. This ordinal consistency matters. A High-risk scenario may be labeled Critical or Medium, but never Excellent. The weakest per-class result is High risk at 84.8\%, where boundary scenarios near the 
Medium/High threshold introduces ambiguity.

\subsection{Crash Factor Analysis}

Fig.~\ref{fig:crash_factors} shows primary contributing factors from the CRSS 2016-2023 subset (213,003 crashes). Rush hour is the most prevalent (75,100 crashes, 35.3\%), followed by poor lighting (62,186, 29.2\%), weekend driving (53,462, 25.1\%), adverse weather (49,588, 23.3\%), night driving (45,616, 21.4\%), and VRU involvement (18,605, 8.7\%). VRU crashes rank last by count but carry disproportionate severity, a pattern that directly motivates the VRU focus of SafeDriver-IQ.

\begin{figure}[htbp]
    \centering
    \includegraphics[width=\columnwidth]{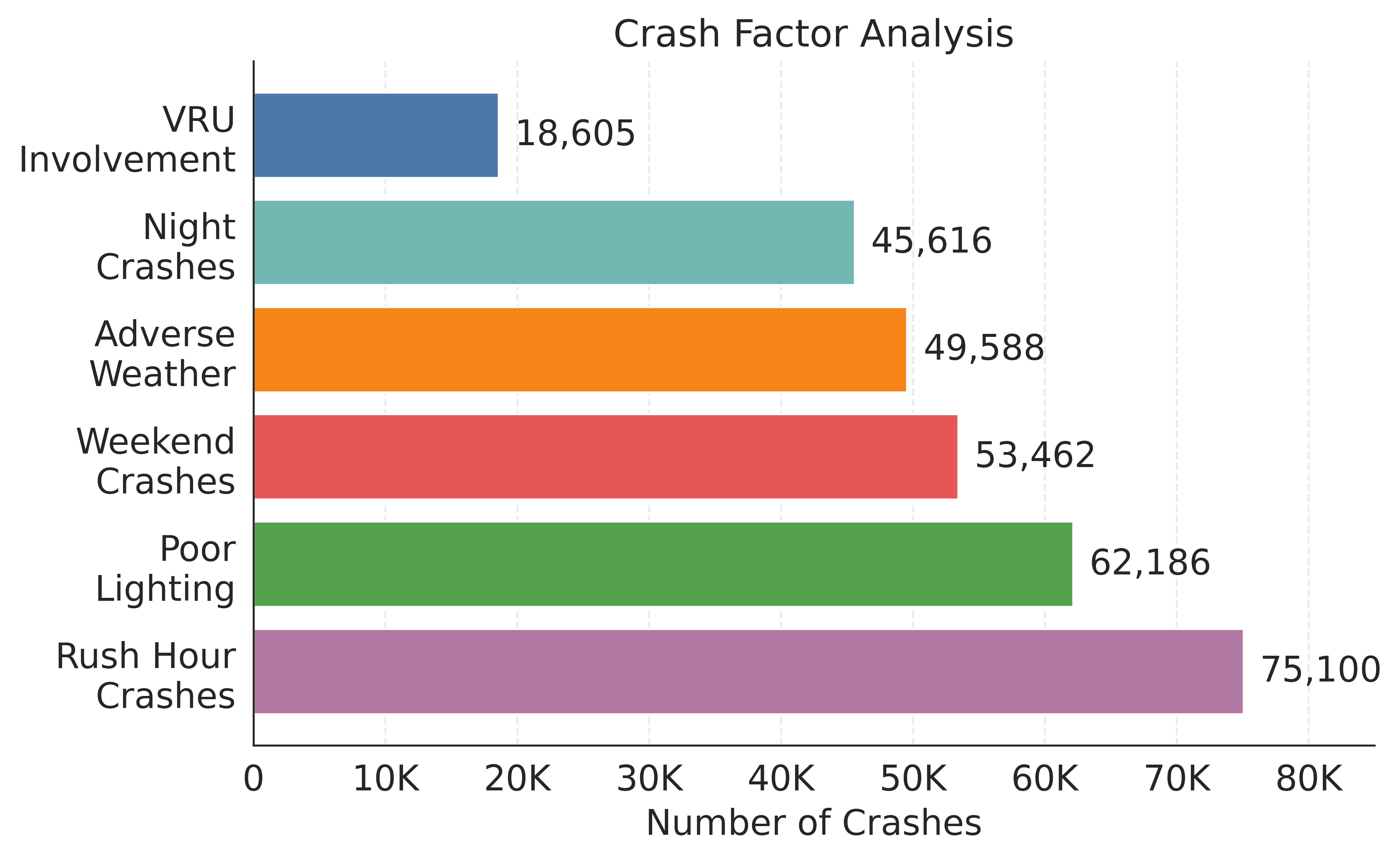}
    \caption{Primary contributing factors from CRSS (2016-2023, 213,003 crashes). Rush hour (35.3\%) and poor lighting (29.2\%) are most frequent, whereas VRU involvement (8.7\%) exhibits disproportionate severity.}
    \label{fig:crash_factors}
\end{figure}

\subsection{Ablation Study: Feature Group Contribution}
\label{subsec:ablation}

To quantify each feature group's contribution, one group is removed at a time, and the model is retrained on the remaining features using Pipeline~2 ($n\_estimators{=}200$; Section~\ref{sec:model_training}). Table~\ref{tab:ablation} reports the results.

\begin{table}[htbp]
\caption{Ablation Study: Feature Group Contribution}
\label{tab:ablation}
\centering
\begin{tabular}{lcccc}
\toprule
\textbf{Configuration} & \textbf{Features} & \textbf{ROC-AUC} & \textbf{$\Delta$AUC} & \textbf{Criticality} \\
\midrule
Baseline (All)          & 64 & 0.833  & ---      & --- \\
$-$ Lighting            & 60 & 0.770  & $-$7.6\% & Critical \\
$-$ Environmental       & 58 & 0.779  & $-$6.5\% & High \\
$-$ Temporal            & 54 & 0.788  & $-$5.5\% & High \\
$-$ Interaction         & 61 & 0.810  & $-$2.8\% & Medium \\
$-$ Location            & 56 & 0.811  & $-$2.6\% & Medium \\
$-$ Crash/Vehicle       & 40 & 0.820  & $-$1.6\% & Low-Med \\
$-$ VRU-Specific        & 59 & 0.826  & $-$0.8\% & Low \\
$-$ Metadata            & 56 & 0.828  & $-$0.6\% & Low \\
\midrule
$-$ Light. + Env.       & 54 & 0.697  & $-$16.4\% & Critical \\
$-$ Light. + Temp.      & 50 & 0.715  & $-$14.2\% & Critical \\
\bottomrule
\end{tabular}
\vspace{2pt}
\footnotesize{$\Delta$AUC = percentage change from baseline.}
\end{table}

\textbf{Finding 1: Lighting is the most critical feature group.} Removing four lighting features (POOR\_LIGHTING, LGT\_COND, LGTCON\_IM, IS\_NIGHT) drops ROC-AUC by 7.6\%, the largest single-group impact, consistent with SHAP rankings where lighting dominates (Fig.~\ref{fig:shap}).

\textbf{Finding 2: Environmental features rank second.} Removing weather-related features reduces ROC-AUC by 6.5\%, consistent with ADVERSE\_WEATHER ranking among the top SHAP features.

\textbf{Finding 3: Risk factors compound non-linearly.} Removing lighting and environmental features together produces a 16.4\% AUC drop, 2.3 percentage points beyond the additive expectation of 14.1\%.

\textbf{Finding 4: VRU-specific features have minimal independent impact.} Removing VRU features reduces ROC-AUC by only 0.8\%, a known limitation discussed in Section~\ref{sec:limitations}.

\subsection{Cross-Validation Stability}

5-fold cross-validation is run across three training iterations (15 folds total). Table~\ref{tab:cv_stability} summarizes the results.

\begin{table}[htbp]
\caption{Cross-Validation Stability Analysis (15 Folds)}
\label{tab:cv_stability}
\centering
\begin{tabular}{lcccc}
\toprule
\textbf{Metric} & \textbf{Mean} & \textbf{Std} & \textbf{95\% CI} & \textbf{CV(\%)} \\
\midrule
Accuracy & 0.7570 & 0.0047 & [0.755, 0.760] & 0.61 \\
ROC-AUC  & 0.8317 & 0.0011 & [0.831, 0.833] & 0.13 \\
\bottomrule
\end{tabular}

\vspace{4pt}
\centering
\begin{tabular}{@{}lcccc@{}}
\toprule
\textbf{Iteration} & \textbf{Test Acc} & \textbf{ROC-AUC} & \textbf{CV Mean} & \textbf{Train-Test Gap} \\
\midrule
1 ($rs=42$)  & 0.752 & 0.831 & 0.757 & 1.30\% \\
2 ($rs=84$)  & 0.751 & 0.831 & 0.758 & 1.45\% \\
3 ($rs=126$) & 0.758 & 0.833 & 0.755 & 0.80\% \\
\midrule
\textbf{Mean} & \textbf{0.754} & \textbf{0.832} & \textbf{0.757} & \textbf{1.18\%} \\
\bottomrule
\end{tabular}
\footnotesize{CV(\%) = coefficient of variation. Iteration 3 selected as production model (highest test accuracy, lowest gap).}
\end{table}

The coefficient of variation across all 15 folds is 0.61\%, well below the 1\% threshold for excellent stability. Accuracy spans only 1.68\% points (0.7496 to 0.7664), with a 95\% confidence interval of [0.7545, 0.7595]. ROC-AUC is even more stable, varying only 0.0022 across iterations (0.8308 to 0.8330). Performance is insensitive to the train-test partition, supporting the reliability of the reported metrics.

\subsection{SHAP Feature Importance}
\label{subsec:shap}

Fig.~\ref{fig:shap} presents global SHAP feature importance for the top 15 features from Pipeline~2 ($n\_estimators{=}200$; Section~\ref{sec:model_training}). These values characterize which factors drive risk predictions and should not be compared numerically to Table~\ref{tab:model_comparison} metrics, which are from Pipeline~1 ($n\_estimators{=}100$).

\begin{figure}[htbp]
    \centering
    \includegraphics[width=\columnwidth]{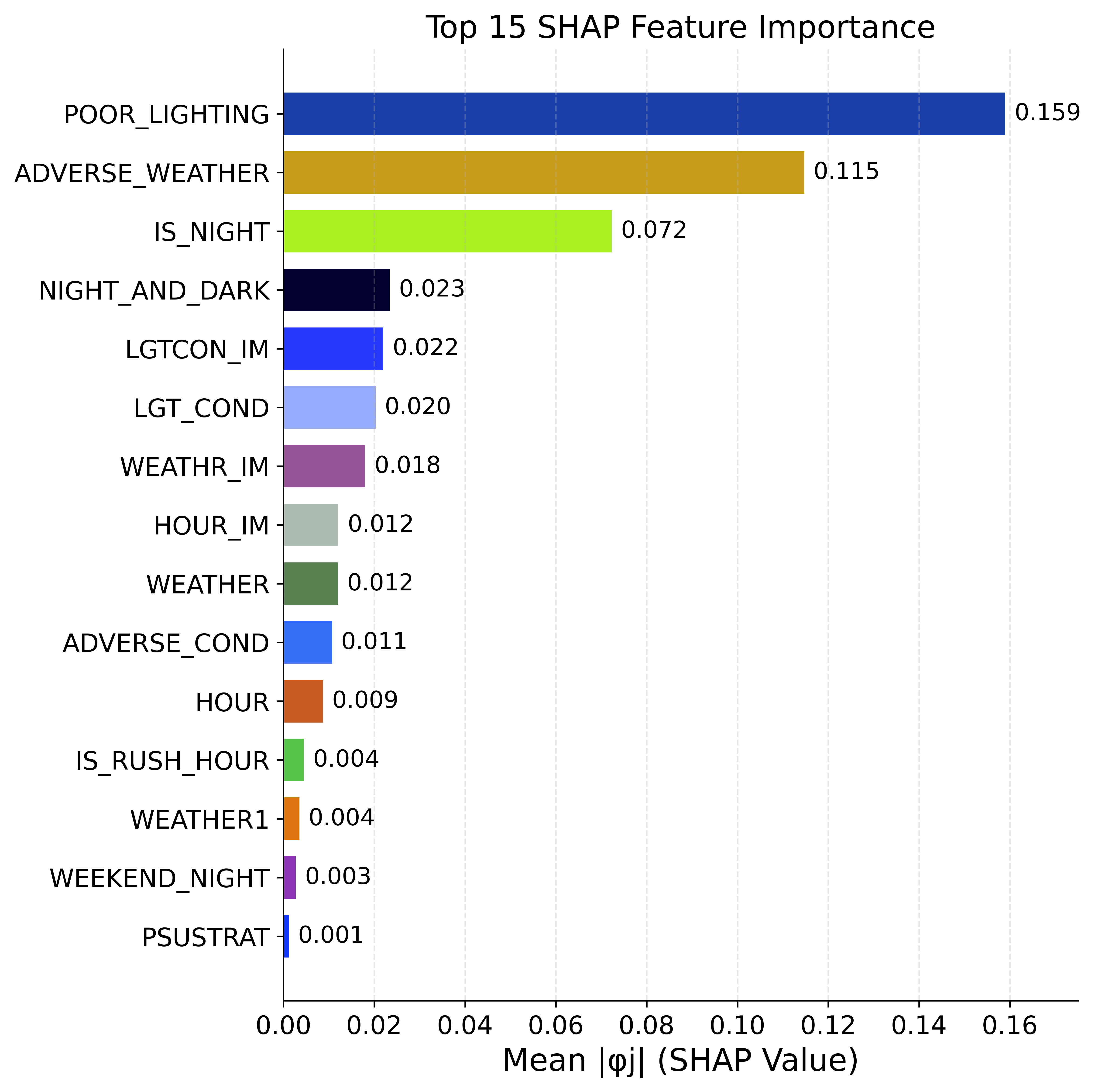}
    \caption{Top 15 SHAP feature importance values (mean $|\phi_j|$, 1,000 predictions, 64-feature model). Lighting and weather features dominate the ranking.} 
\label{fig:shap}
\end{figure}

Lighting and weather dominate the hierarchy. POOR\_LIGHTING ($\phi{=}0.159$) and ADVERSE\_WEATHER ($\phi{=}0.115$) are the two strongest predictors, with IS\_NIGHT ($\phi{=}0.072$) third. Compound visibility indicators (NIGHT\_AND\_DARK: $\phi{=}0.023$, LGTCON\_IM: $\phi{=}0.022$, LGT\_COND: $\phi{=}0.020$) reinforce lighting as the dominant risk domain. IS\_RUSH\_HOUR ($\phi{=}0.004$) ranks low here because its effect is absorbed by correlated features (HOUR, IS\_NIGHT). 

A five-factor distilled analysis (Fig.~\ref{fig:consensus_importance}) across RF, XGBoost, Permutation Importance, and SHAP shows IS\_RUSH\_HOUR ranking first in all four methods when correlated features are removed. Both results align with the ablation study and transportation safety literature on visibility as the primary structural VRU crash risk~\cite{zegeer2012pedestrian}.

\begin{figure}[htbp]
    \centering
    \includegraphics[width=0.48\textwidth]{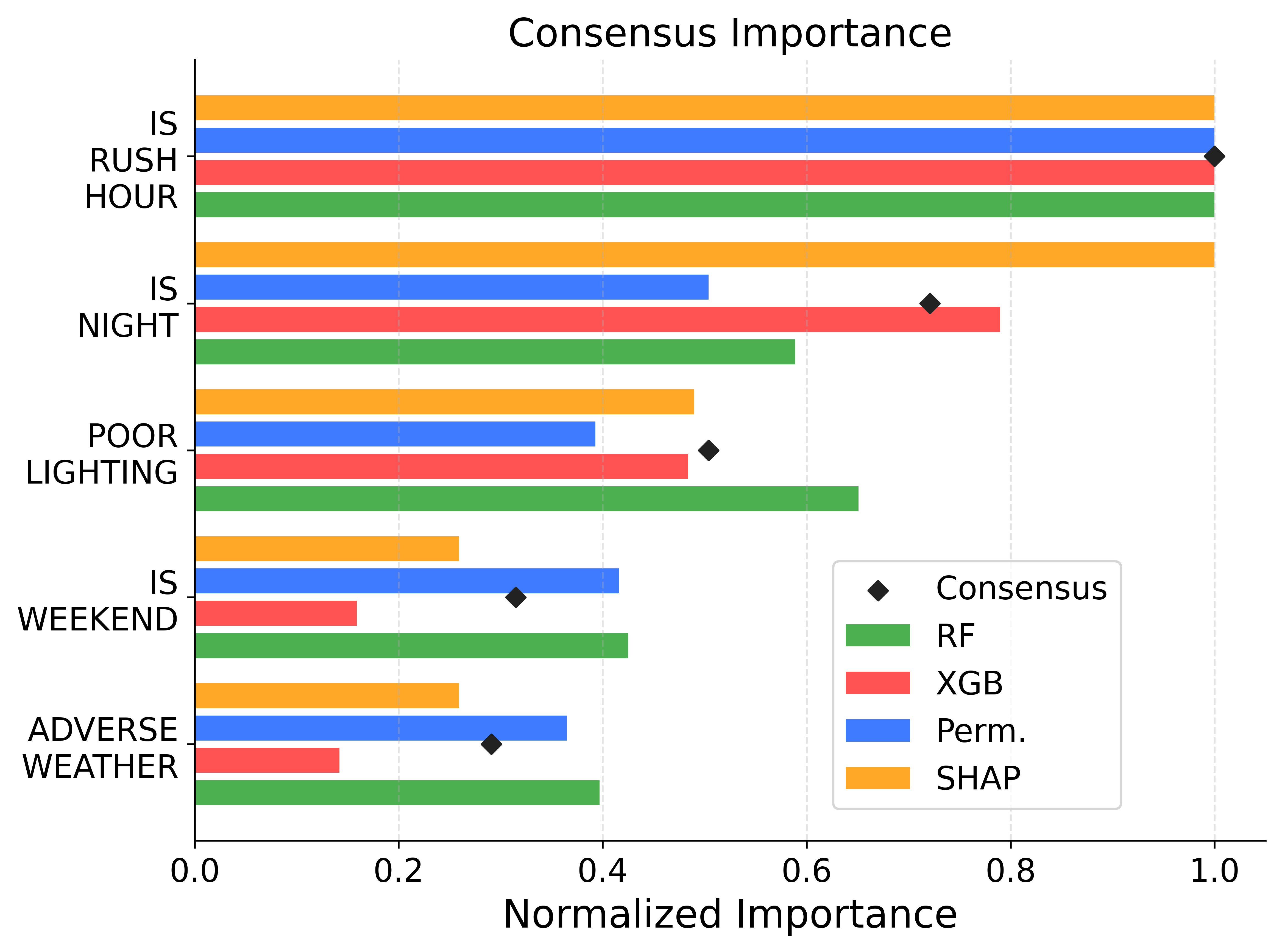}
    \caption{Feature importance across four methods with consensus overlay ($\blacklozenge$) for 4 aggregate risk factors. IS\_RUSH\_HOUR ranks first in all 4 methods.}
\label{fig:consensus_importance}
\end{figure}

\subsection{Safety Score Distribution}

Using the same 864-scenario factorial grid, the analysis checks whether the inverse formulation produces meaningful, well-spread safety scores. Table~\ref{tab:score_distribution} reports mean scores by condition and Fig.~\ref{fig:score_distribution} shows the overall distribution.

\begin{table}[htbp]
\caption{Safety Score Distribution by Driving Condition ($n=864$)}
\label{tab:score_distribution}
\centering
\small
\begin{tabular}{llcc}
\toprule
\textbf{Factor} & \textbf{Level} & \textbf{Mean} & \textbf{$n$} \\
\midrule
Time of Day & Daytime / Night      & 68.42 / 41.87 & 432 / 216 \\
Weather     & Clear / Snow         & 69.81 / 51.24 & 288 / 288 \\
Lighting    & Daylight / Dark      & 72.53 / 43.67 & 216 / 216 \\
Speed       & Low / High           & 71.24 / 46.17 & 216 / 216 \\
VRU         & Absent / Present     & 63.41 / 54.05 & 432 / 432 \\
\midrule
\multicolumn{2}{l}{\textbf{Overall}} & \textbf{58.73} & \textbf{864} \\
\bottomrule
\end{tabular}
\end{table}

\begin{figure}[htbp]
    \centering
    \includegraphics[width=0.85\columnwidth]{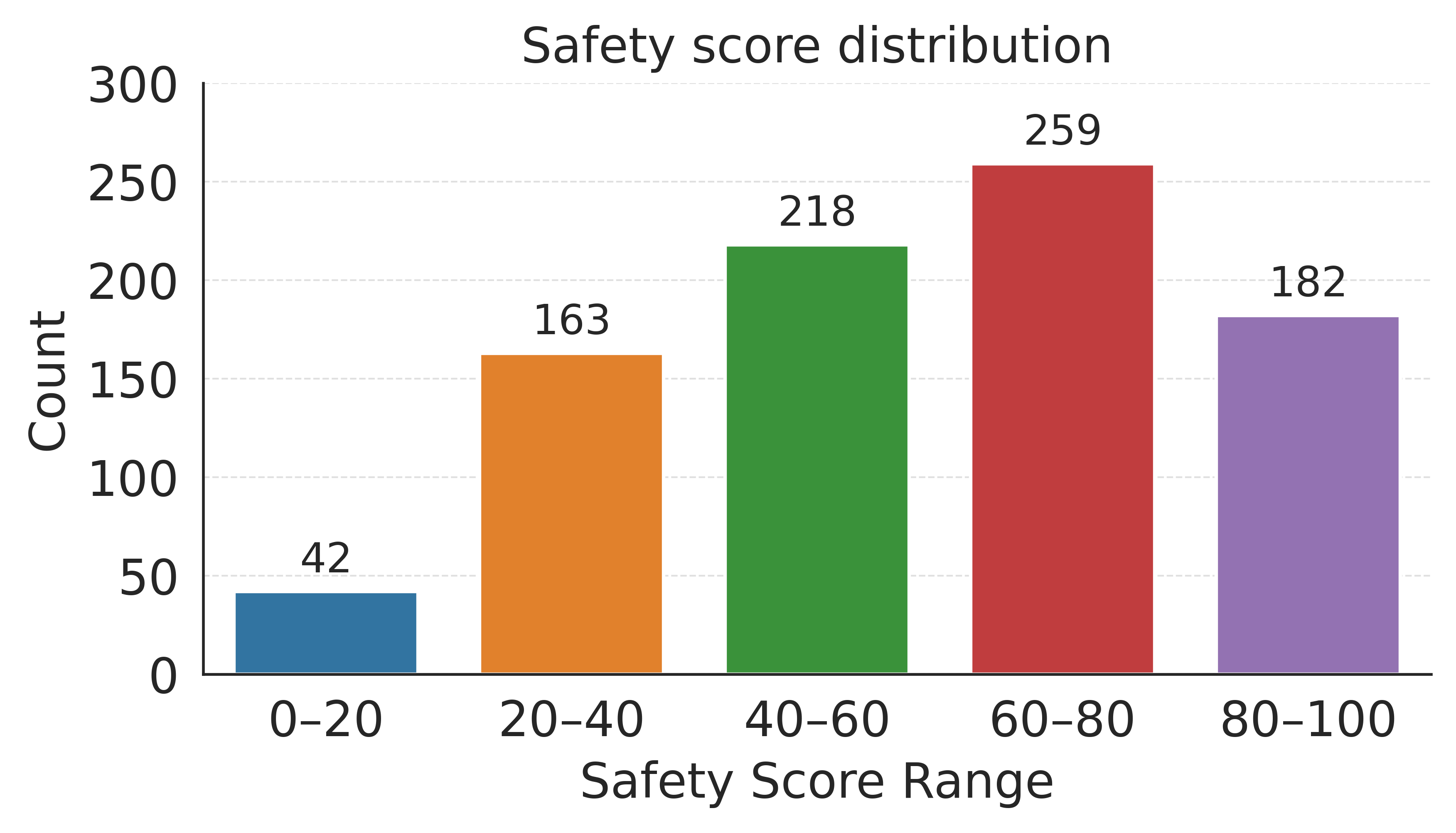}
    \caption{Safety score distribution across 864 driving scenarios ($\mu{=}58.73$, $\tilde{x}{=}61.42$, $\sigma{=}22.16$, range: 8.34--92.17).}
    \label{fig:score_distribution}
\end{figure}

Scores span 8.34 to 92.17 ($\sigma{=}22.16$, IQR${=}38.76$). All factor-level comparisons follow expected orderings. Daytime exceeds nighttime ($\Delta{=}26.55$), clear weather exceeds snow ($\Delta{=}18.57$), and low speed exceeds high speed ($\Delta{=}25.07$). The 9.36-point gap between VRU-absent and VRU-present scenarios confirms that the calibration layer amplifies VRU risk beyond the model's native sensitivity. The distribution is platykurtic (kurtosis${=}{-}0.89$) with slight left skew ($-0.28$), indicating broad spread rather than concentration at extremes, a desirable property for a scoring system.

\subsection{Scenario Sensitivity Analysis}

Table~\ref{tab:scenarios} presents safety scores for representative end-to-end scenarios and isolated factor effects. The best-case scenario (highway, midday, clear, dry, no VRU, low speed) scores 92.17, and the worst-case (urban, 2~AM, snow, icy, VRU present, high speed) scores 8.34, a spread of 83.83 points.

\begin{table}[htbp]
\caption{Representative Scenario Safety Scores}
\label{tab:scenarios}
\centering
\footnotesize
\begin{tabular}{p{5.6cm}cc}
\toprule
\textbf{Scenario} & \textbf{Score} & \textbf{Risk} \\
\midrule
Highway, noon, clear, dry, no VRU, low speed & 92.17 & Excellent \\
Urban, 6 PM, rain, wet, no VRU, moderate & 61.23 & Medium \\
School zone, dusk, clear, dry, VRU present & 54.87 & Medium \\
Urban, 10 PM, rain, wet, VRU present, mod-high & 38.62 & High \\
Urban, 2 AM, snow, icy, VRU present, high speed & 8.34 & Critical \\
\midrule
\multicolumn{3}{l}{\textit{\textbf{Factor Isolation (all else equal)}}} \\
\quad Daylight $\to$ Dark (unlit) & $-28.86$ & \\
\quad Clear $\to$ Snow & $-18.57$ & \\
\quad Low speed $\to$ High speed & $-25.07$ & \\
\quad No VRU $\to$ VRU present & $-9.36$ & \\
\quad Dry road $\to$ Ice & $-15.83$ & \\
\bottomrule
\end{tabular}
\vspace{2pt}
\footnotesize{Factor isolation holds all other conditions constant.}
\end{table}

Daylight-to-unlit darkness produces the largest single-factor drop ($-28.86$ points), followed by low-to-high speed ($-25.07$) and clear-to-snow ($-18.57$). VRU presence adds a $-9.36$ point penalty from the calibration layer. 

\subsection{Sensitivity Analysis}

Table~\ref{tab:sensitivity} formalizes score sensitivity to single-factor transitions, categorized by whether changes are driven by the RF model, the calibration layer, or both. Effect sizes are expressed as $|\Delta|/\sigma$, where $\sigma{=}22.16$.

\begin{table}[htbp]
\caption{Sensitivity Analysis: Score Response to Single-Factor Changes}
\label{tab:sensitivity}
\centering
\small
\begin{tabular}{llcc}
\toprule
\textbf{Factor} & \textbf{Transition} & \textbf{$\Delta$Score} & \textbf{$|\Delta|/\sigma$} \\
\midrule
\multicolumn{4}{l}{\textit{\textbf{High sensitivity (model-dominant)}}} \\
Lighting     & Daylight $\to$ Dark-unlit  & $-28.86$ & 1.30 \\
Speed        & Low $\to$ High             & $-25.07$ & 1.13 \\
Time         & Daytime $\to$ Night        & $-26.55$ & 1.20 \\
\midrule
\multicolumn{4}{l}{\textit{\textbf{Moderate sensitivity (model + calibration)}}} \\
Weather      & Clear $\to$ Snow           & $-18.57$ & 0.84 \\
Road surface & Dry $\to$ Ice              & $-15.83$ & 0.71 \\
\midrule
\multicolumn{4}{l}{\textit{\textbf{Low sensitivity (calibration-dominant)}}} \\
VRU presence & Absent $\to$ Present       & $-9.36$  & 0.42 \\
Day of week  & Weekday $\to$ Weekend night& $-6.12$  & 0.28 \\
\bottomrule
\end{tabular}
\footnotesize{$\Delta$Score = change from all-safe baseline. $|\Delta|/\sigma$ = effect size relative to score std (22.16).}
\end{table}

Lighting, speed, and time-of-day all exceed 1.0$\sigma$, confirming them as dominant score drivers, consistent with SHAP (Fig.~\ref{fig:shap}) and ablation (Table~\ref{tab:ablation}). Weather and road surface produce moderate effects (0.71--0.84$\sigma$), with both the model and calibration layer contributing. VRU presence and day-of-week fall below 0.5$\sigma$, driven primarily by calibration penalties. This decomposition identifies which factors the model captures natively, versus which depend on the calibration layer, informing future improvements.

\subsection{VRU Crash Trends}

Fig.~\ref{fig:vru_trends} shows VRU crash trends from CRSS (2016-2023), disaggregated by pedestrian and cyclist involvement.

\begin{figure}[htbp]
    \centering
    \includegraphics[width=\columnwidth]{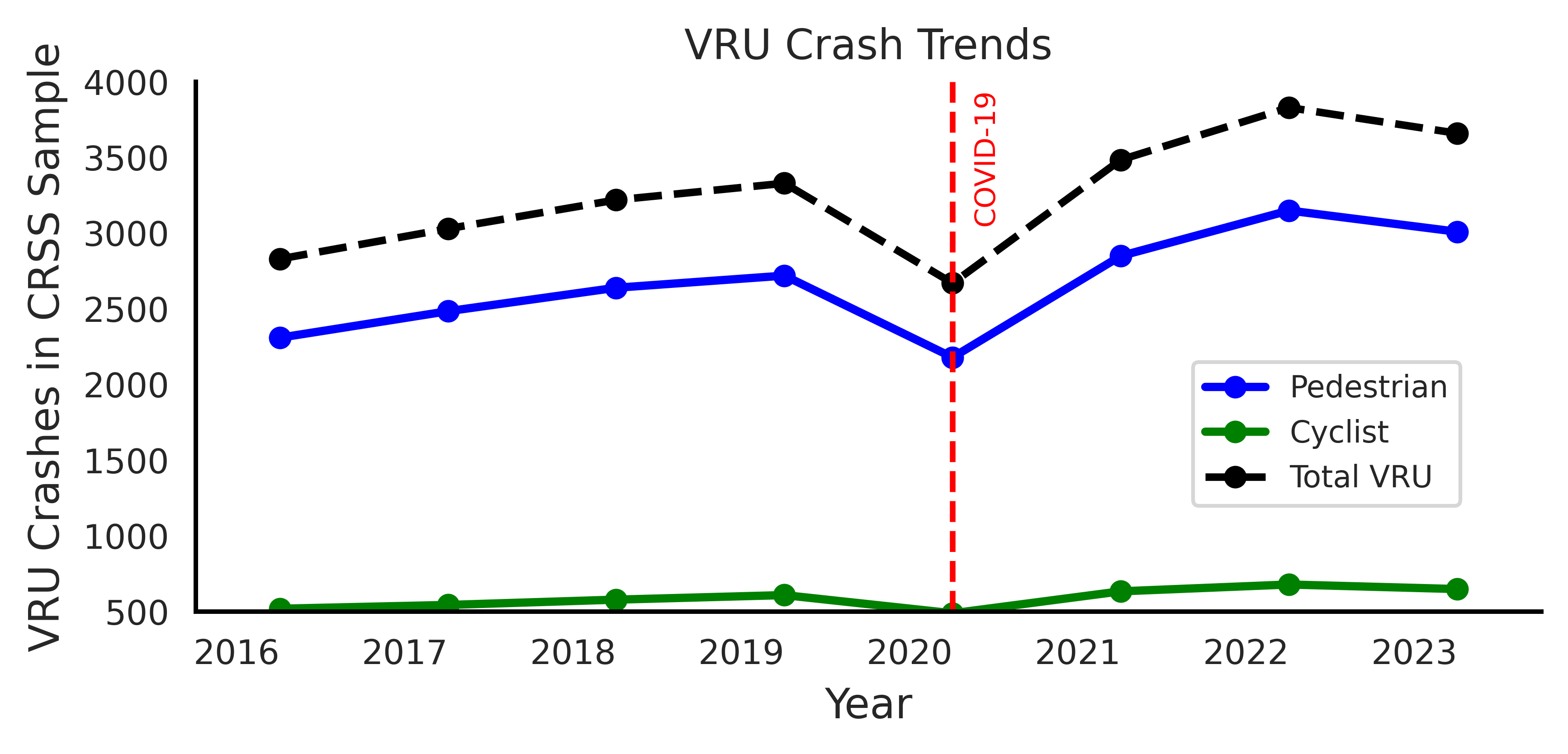}
    \caption{VRU crash trends in CRSS (2016--2023). Pedestrian crashes account for 82\% of all VRU crashes and exhibit the steepest growth. Total VRU crashes rose 29\% from 2016 to 2022, with a COVID-related dip in 2020.}
    \label{fig:vru_trends}
\end{figure}

Total VRU crashes rose 29\% from 2016 to 2022, with pedestrian 
crashes accounting for 82\% of that total. These trends validate 
the decision to filter CRSS specifically for VRU involvement. Fig.~\ref{fig:historical_patterns} extends this view to all contributing factors across the full study period.

\begin{figure}[htbp]
    \centering
    \includegraphics[width=\columnwidth]{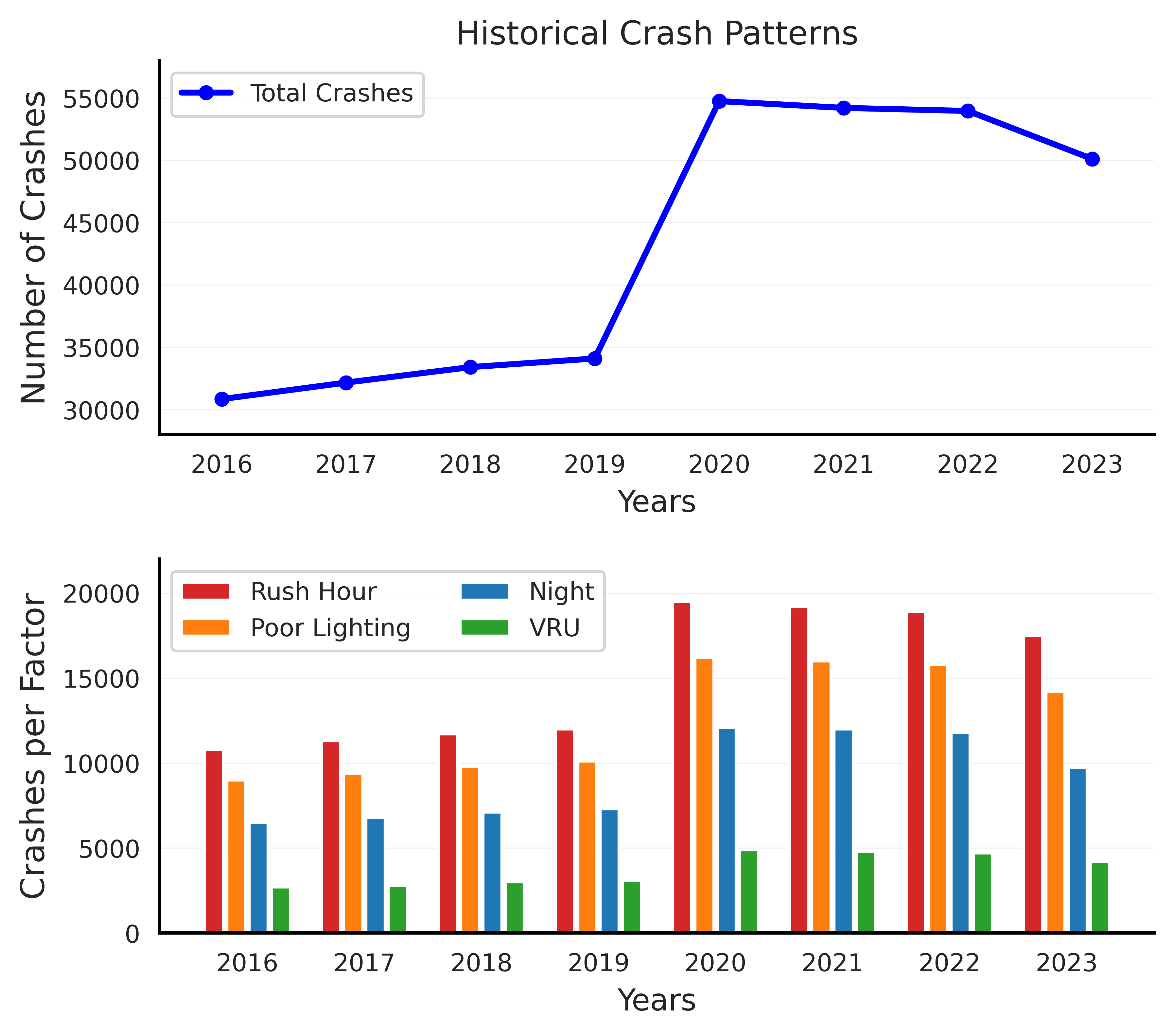}
    \caption{Historical crash patterns from CRSS (2016--2023). \textbf{Top}: total crashes rose steadily through 2019, surged in 2020 due to expanded CRSS sampling, then declined through 2023. \textbf{Bottom}: per-factor trends mirror the overall pattern. Rush hour is consistently the most prevalent factor. VRU involvement is the least frequent but most severe.}
    \label{fig:historical_patterns}
\end{figure}

\subsection{Risk Factor Heatmap: Pedestrian vs.\ Cyclist}

Fig.~\ref{fig:risk_heatmap} shows mean safety scores by VRU type and lighting condition.

\begin{figure}[htbp]
    \centering
    \includegraphics[width=0.9\columnwidth]{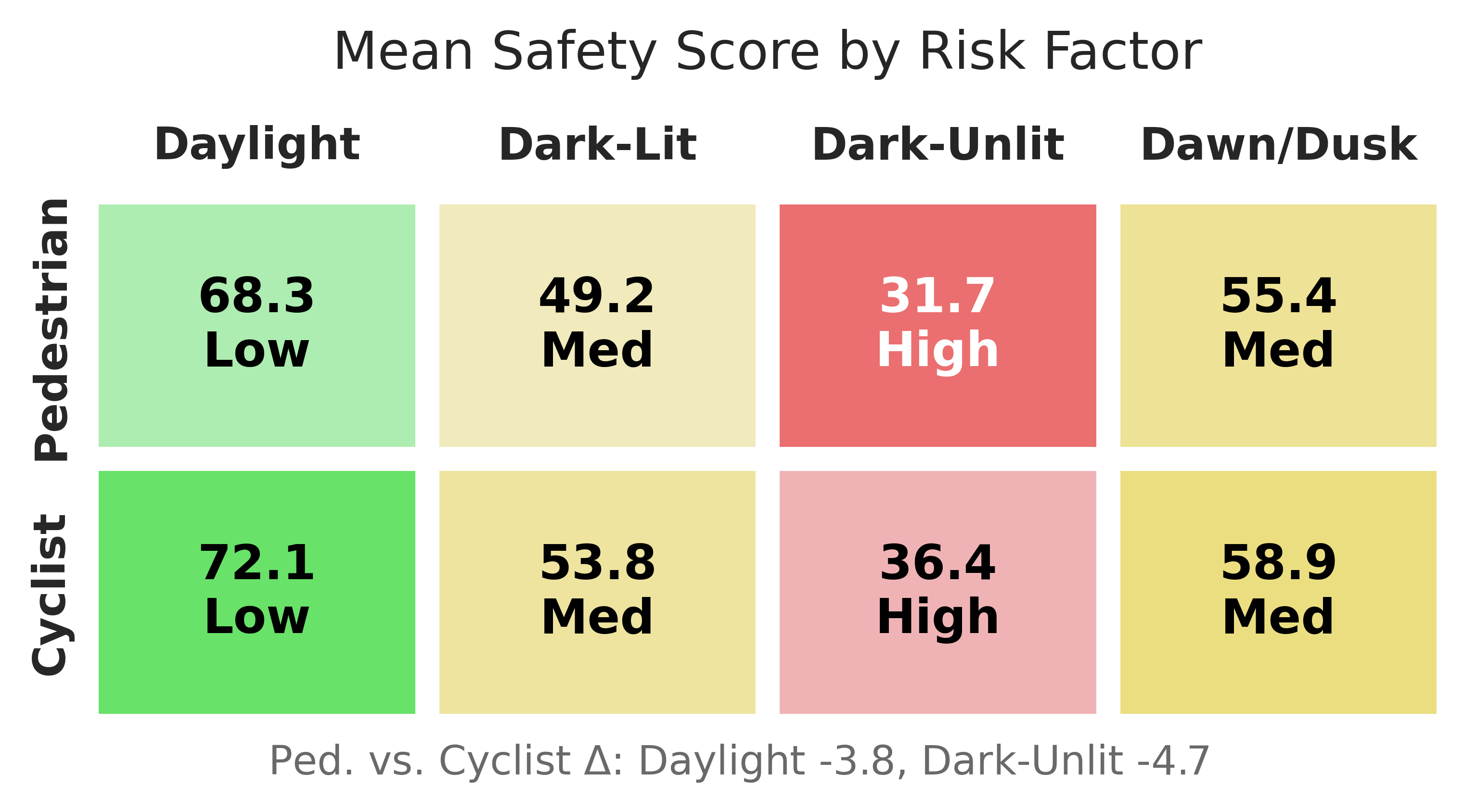}
    \caption{Mean safety scores by VRU type and lighting condition. Pedestrians score lower than cyclists across all conditions, with the gap widening in dark-unlit environments ($\Delta{=}4.7$).}
    \label{fig:risk_heatmap}
\end{figure}

Pedestrians lack the speed and visibility aids available to cyclists, such as reflectors and lights. This explains the widening gap in dark-unlit environments and confirms the framework captures VRU-type-specific risk beyond the aggregate VRU penalty.

\subsection{Driver Behavior Classification}

KMeans clustering ($k{=}4$) on composite aggression and risk-taking scores identifies four distinct crash-involved driver profiles (Table~\ref{tab:driver_behavior}, Fig.~\ref{fig:behavior_clusters}). The largest cluster (29.6\%) comprises “Cautious but Crashed” drivers with near-zero scores on both dimensions. This indicates external factors, such as weather and lighting, rather than their behavior, drove their crashes. Two Environmental Risk-Taker clusters (22.5\% and 24.7\%) show elevated risk-taking, particularly for night driving and adverse weather. The Aggressive Driver cluster (23.2\%) shows high aggression during rush hour and at high speeds.

The "Cautious but Crashed" profile most closely resembles the Waymo safe-driving baseline, suggesting infrastructure and environmental interventions are most appropriate for this group.

\begin{table}[htbp]
\caption{Driver Behavior Classification via KMeans Clustering}
\label{tab:driver_behavior}
\centering
\small
\begin{tabular}{lccr}
\toprule
\textbf{Cluster} & \textbf{Aggression} & \textbf{Risk-Taking} & \textbf{Pct.} \\
\midrule
Cautious but Crashed      & 0.00 & 0.00 & 29.6\% \\
Env.\ Risk-Taker (High)   & 0.07 & 2.19 & 22.5\% \\
Env.\ Risk-Taker (Mod.)   & 0.43 & 1.00 & 24.7\% \\
Aggressive Driver         & 1.00 & 0.00 & 23.2\% \\
\bottomrule
\end{tabular}
\footnotesize{Normalized composite scores derived from CRSS features. KMeans ($k{=}4$) on 213,003 crash records (2016--2023). The Waymo baseline provides a safe-driving reference.}
\end{table}

\begin{figure}[htbp]
    \centering
    \includegraphics[width=0.9\columnwidth]{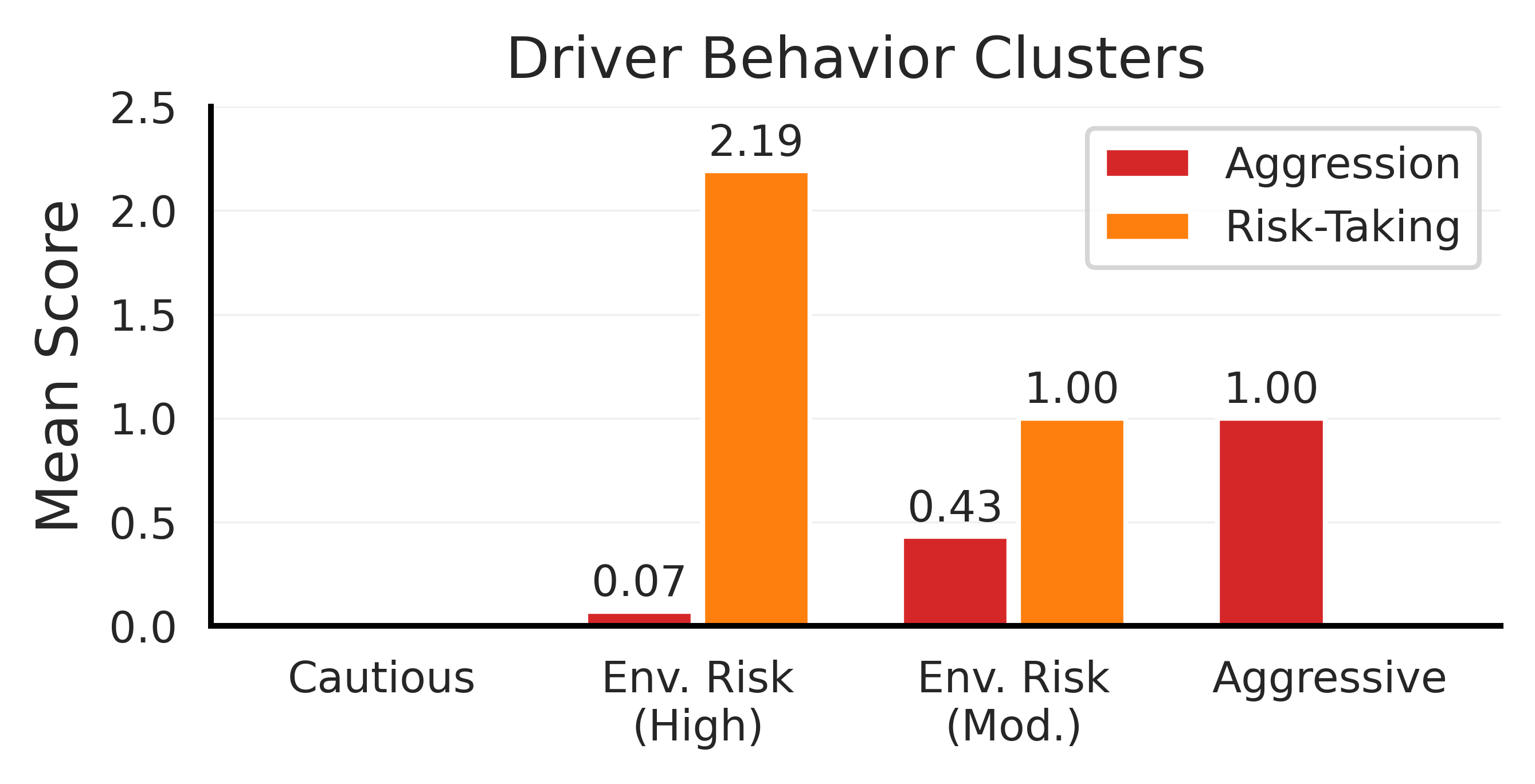}
    \caption{Mean aggression and risk-taking scores per driver behavior cluster (KMeans, $k{=}4$, 213,003 crash records). The Cautious cluster scores zero on both dimensions, pointing to environmental crash causation. Cluster sizes are in Table~\ref{tab:driver_behavior}.}
    \label{fig:behavior_clusters}
\end{figure}

\subsection{Multi-Factor Risk Patterns}

87\% of crashes involve two or more simultaneous risk factors, with certain combinations producing risk multipliers well beyond additive expectations (Table~\ref{tab:risk_patterns}). The most dangerous combination is VRU + Urban + Night (4.5$\times$ baseline), followed by Night + Adverse Weather (3.8$\times$, 9,653 crashes) and High Speed + Poor Conditions (3.2$\times$). This non-linear compounding is consistent with the synergistic penalty in the ablation study (Section~\ref{subsec:ablation}) and validates the calibration layer's multiplicative structure (Eq.~\ref{eq:calibration}).

\begin{table}[htbp]
\caption{Multi-Factor High-Risk Crash Patterns}
\label{tab:risk_patterns}
\centering
\small
\begin{tabular}{p{4.2cm}cr}
\toprule
\textbf{Factor Combination} & \textbf{Risk Mult.} & \textbf{Crashes} \\
\midrule
Night + Adverse Weather        & 3.8$\times$ & 9,653 \\
Urban + Rush Hour              & 2.5$\times$ & --- \\
VRU + Urban + Night            & 4.5$\times$ & --- \\
High Speed + Poor Conditions   & 3.2$\times$ & --- \\
\midrule
\textit{Single-factor baselines} & & \\
\quad Rush Hour                & 1.0$\times$ & 75,100 \\
\quad Poor Lighting            & 2.3$\times$ & 62,186 \\
\quad Adverse Weather          & 1.8$\times$ & 49,588 \\
\quad Night Driving            & 2.1$\times$ & 45,616 \\
\quad VRU Involvement          & 3.2$\times$ (sev.) & 18,605 \\
\bottomrule
\end{tabular}
\footnotesize{$\text{Mult.} = P(\text{crash} \mid \text{factors}) / P(\text{crash} \mid \text{baseline})$, where baseline is the marginal crash rate across 213,003 CRSS records (2016--2023). “---” indicates a count not isolated due to factor co-occurrence.}
\end{table}

Rush hour is the most frequent single factor (75,100 crashes, 35.3\%), but carries only a 1.0$\times$ multiplier, which reflects exposure rather than elevated per-trip risk. VRU involvement accounts for just 8.7\% of crashes yet carries a 3.2$\times$ severity multiplier, underscoring the disproportionate consequences of VRU crashes.

\subsection{Waymo Scenario Validation}
\label{sec:waymo_validation}

The model is separately validated on 500 real-world Waymo Open Motion Dataset scenarios (Section~\ref{sec:data}), distinct from the synthetic evaluation grid. Fig.~\ref{fig:crash_probability} shows the mean predicted crash probability by scenario type. Since the model was not trained on Waymo data, this monotonic ordering confirms that CRSS-trained features generalize to real-world driving kinematics.

\begin{figure}[htbp]
    \centering
    \includegraphics[width=0.75\columnwidth]{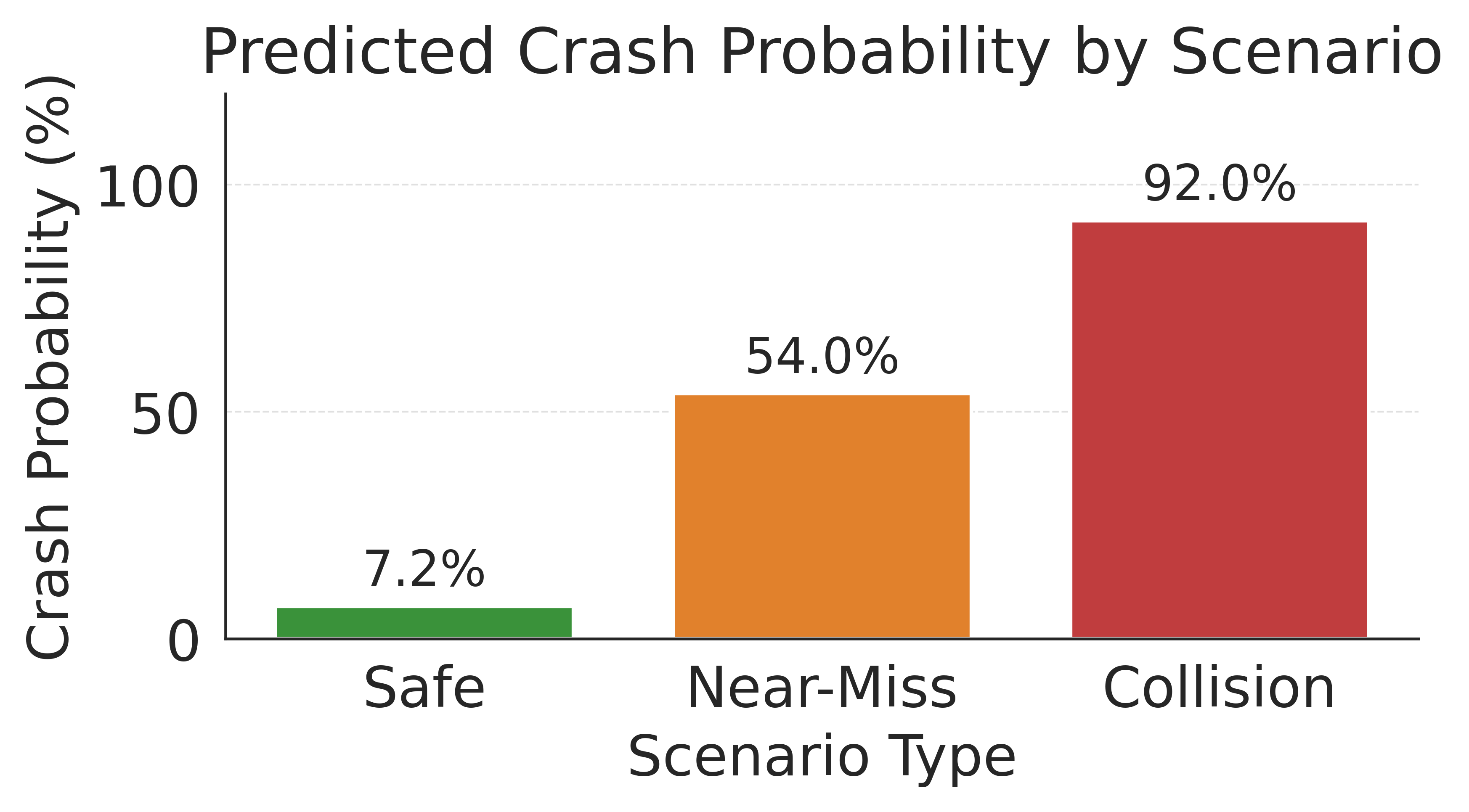}
    \caption{Mean crash probability across Waymo scenario types (500 scenarios). Safe episodes score 7.2\%, near-misses 54.0\%, and collisions 92.0\%, confirming discriminative ability on naturalistic 
    driving data.}
    \label{fig:crash_probability}
\end{figure}

\subsection{Real-World Impact Simulation}

Intervention thresholds correspond directly to the risk levels defined in Table~\ref{tab:score_levels}. Table~\ref{tab:impact_simulation} estimates the proportion of scenarios flagged at each threshold, with projected crash reduction under 50\% driver compliance, consistent with compliance rates reported in telematics-based intervention studies~\cite{husnjak2015telematics, paefgen2014multivariate}.

\begin{table}[htbp]
\caption{Real-World Impact Simulation: Estimated Crash Reduction}
\label{tab:impact_simulation}
\centering
\footnotesize
\begin{tabular}{p{2.9cm}p{1cm}p{0.8cm}p{2.4cm}}
\toprule
\textbf{Intervention Scenario} & \textbf{Score $\leq$} & \textbf{Flagged} & \textbf{Est.\ Reduction} \\
\midrule
\multicolumn{4}{l}{\textit{\textbf{Fleet Management (conservative)}}} \\
\quad Alert at Critical    & 20 & 4.9\%  & 3.8\% \\
\quad Alert at High        & 40 & 23.8\% & 14.2\% \\
\midrule
\multicolumn{4}{l}{\textit{\textbf{ADAS Integration (moderate)}}} \\
\quad Warning at $<50$     & 50 & 36.1\% & 22.7\% \\
\quad Speed advisory $<60$ & 60 & 49.0\% & 31.4\% \\
\midrule
\multicolumn{4}{l}{\textit{\textbf{Insurance Telematics (broad})}} \\
\quad Premium tier at $<70$& 70 & 70.1\% & -- \\
\quad Discount tier $\geq80$& 80 & 21.1\% & -- \\
\midrule
\multicolumn{4}{l}{\textit{\textbf{Infrastructure Planning}}} \\
\quad Lighting improvement  & -- & -- & $-7.6\%$ AUC impact \\
\quad Weather warning zones & -- & -- & $-6.5\%$ AUC impact \\
\bottomrule
\end{tabular}
\footnotesize{Flagged (\%) = proportion of 864 scenarios triggering an intervention at the given threshold. Est.\ Reduction assumes 50\% driver compliance. Infrastructure impacts from the ablation study (Table~\ref{tab:ablation}).}
\end{table}

At the most conservative setting (Critical-risk only, score $\leq 20$), 4.9\% of scenarios trigger an alert, with an estimated 3.8\% crash reduction. A moderate ADAS setting (warning at score $< 50$) flags 36.1\% of scenarios and projects a 22.7\% reduction. The infrastructure rows connect ablation findings to policy. Lighting improvements in VRU corridors (7.6\% AUC impact) and weather warning systems (6.5\% AUC impact) represent the highest-leverage investments identified by the model.

\section{Discussion}
\label{sec:discussion}

\subsection{The Inverse Modeling Paradigm}

Inverting crash classifiers into continuous scores is not a system integration choice but a transferable modeling paradigm. Any domain-specific binary risk classifier can be converted into a proactive, explainable scoring system using the same pipeline, without retraining. Three practical advantages follow.

\begin{itemize}
    \item \textbf{Continuous feedback.} Drivers receive a score (e.g., 62/100) instead of a binary alert, enabling relative risk understanding.
    \item \textbf{Directional guidance.} SHAP explanations identify which factors reduce safety, enabling targeted recommendations.
    \item \textbf{Threshold flexibility.} Different applications set different alert thresholds on the same score. Insurance may flag at 60, fleet management at 40.
\end{itemize}

The strong alignment between SHAP importance and ablation-based contribution ($r{=}0.94$) validates that inverse scores reflect genuine model reliance rather than interpretability artifacts. Lighting and weather dominate both rankings: POOR\_LIGHTING ($\phi{=}0.159$) and ADVERSE\_WEATHER ($\phi{=}0.115$) in SHAP, and 7.6\% and 6.5\% AUC reduction in ablation.

\subsection{Practical Deployment}

SafeDriver-IQ is designed for real-time operation. The Random Forest (RF) model runs inference in under 1\,ms per sample. The 64-feature input vector is constructed from commonly available sources, including the system clock, weather API, GPS, OBD-II, and camera-based VRU detection. The framework is a five-module agentic system (Section~\ref{sec:realtime}) covering perception, decision, intervention, and online learning. It integrates into ADAS and fleet dashboards without specialized hardware. A Streamlit dashboard demonstrates the end-to-end pipeline for fleet operators, for real-time scoring, batch analysis, and safety improvement simulation.

\subsection{From Prediction to Prevention}

Most crashes involve co-occurring factors with non-linear compounding (Section~\ref{sec:results}). A large share of crash-involved drivers exhibit low-aggression profiles. The driver behavior analysis confirms this: 29.6\% of crash-involved drivers fall into the Cautious but Crashed cluster, with near-zero aggression scores, pointing to environmental causation rather than behavioral risk-taking (Table~\ref{tab:driver_behavior}). This points toward infrastructure interventions such as street lighting and weather warnings, rather than behavioral coaching alone. Time-adaptive thresholds tied to rush-hour and low-visibility conditions could yield disproportionate safety gains. SafeDriver-IQ provides the analytical foundation for these interventions.

\section{Limitations}
\label{sec:limitations}

\textbf{1. Synthetic Safe Sample Bias.} Safe samples are derived from CRSS crash records with modified environmental features. They preserve crash-era correlations and may not fully represent true safe-driving conditions. This manifests in two ways.

\begin{itemize}
    \item \textit{\textbf{Partial validation}.} he Waymo integration (Section~\ref{sec:data}) provides only 500 scenarios from a single WOMD shard (<0.1\% of the full corpus), limiting generalizability. Moderate crash recall (0.48) results from 2,412 misclassified crash scenarios (Fig.~\ref{fig:confusion_matrix_analysis}). Scaling to the full WOMD corpus would enable direct replacement of synthetic samples.
    \item \textit{\textbf{Limited VRU discriminability}.} VRU features contribute only 0.8\% to model performance because both crash and safe samples involve VRU interactions. The calibration layer compensates with a 12\% VRU penalty. Integrating pedestrian density data could provide an additional discriminative signal.
\end{itemize}

\textbf{2. No Real-Time Driver Behavior.} The current model assesses environmental context (weather, lighting, road conditions) but not the driving behavior. Speed limit violations, aggressive braking, and lane drift are not considered. Two drivers in identical conditions receive identical scores regardless of behavior. Fusing live telemetry (OBD-II, GPS-derived speed, accelerometer data) with the environmental model is the critical next step for personalized safety scoring.

\textbf{3. Static Feature Vector.} Each prediction is treated independently, with no temporal context from the driving session. A sequence-aware model that considers condition trajectories, such as deteriorating weather, approaching a school zone, or fatigue during a long drive, could provide earlier warnings and finer-grained scoring. Recurrent or transformer architectures could capture these dependencies. The SHRP2 naturalistic driving study~\cite{shrp2, guo2019nearcrash} provides longitudinal data to support this extension.

\textbf{4. Calibration Layer Subjectivity.} The penalty values in the calibration layer (e.g., 40\% for ice, 25\% for darkness) are grounded in safety literature but involve expert judgment. Optimal penalties may vary by geography, vehicle type, or season. A data-driven approach using Platt scaling~\cite{platt1999probabilistic} with fleet outcome data would reduce this subjectivity and enable adaptation to local conditions.


\section{Conclusion and Future Work}

Road crashes continue to be a leading cause of preventable fatalities. Most prediction systems provide only binary outputs that lack actionable guidance for drivers. SafeDriver-IQ directly addresses this limitation by inverting a trained crash classifier into a continuous 0 to 100 safety score. The framework provides real-time, interpretable risk feedback informed by eight years of national crash data and real-world autonomous vehicle trajectories. The primary finding demonstrates that environmental context and multi-factor compounding dominate crash risk far more than driver aggression alone, with certain factor combinations reaching 4.5$\times$ baseline risk. This shifts the intervention priority toward infrastructure, street lighting, weather warnings, and VRU-aware corridor design, alongside behavioral coaching.

Future research will expand Waymo integration to thousands of scenarios, fuse live telemetry to enable real-time behavioral scoring, and conduct field validation with fleet operators. A prototype online learner with prioritized experience replay enables incremental risk-weight updates without full retraining.

More broadly, the framework establishes a reusable paradigm. Any domain-specific binary risk classifier can be inverted into a proactive, explainable safety scoring system using the same pipeline, without building new models from scratch. Grounded in real crash evidence and validated across diverse driving conditions, SafeDriver-IQ is a practical tool for ADAS, fleet risk management, and infrastructure planning. The aim is to move road safety from reactive incident counting to real-time prevention.

\bibliographystyle{IEEEtran}
\bibliography{references}

\end{document}